\documentclass{article}

\usepackage[preprint]{neurips_2026}

\usepackage[utf8]{inputenc} 
\usepackage[T1]{fontenc}    
\usepackage{hyperref}       
\usepackage{url}            
\usepackage{booktabs}       
\usepackage{amsfonts}       
\usepackage{nicefrac}       
\usepackage{microtype}      
\usepackage{xcolor}         
\usepackage{amsmath} 
\usepackage{bbm}
\usepackage{graphicx}
\usepackage{amssymb} 
\usepackage{tcolorbox}
\tcbuselibrary{skins,breakable}

\newcounter{takeawaycounter}
\newtcolorbox{takeaway}{
  enhanced, breakable,
  colback=blue!5!white,
  colframe=blue!70!black,
  fonttitle=\bfseries\footnotesize,
  before upper={\small},
  left=6pt, right=6pt, top=6pt, bottom=4pt,
  arc=2pt,
  boxrule=0.6pt,
  attach boxed title to top left={yshift=-2mm, xshift=4mm},
  boxed title style={
    colback=blue!70!black,
    colframe=blue!70!black,
    coltext=white,
    arc=2pt,
    boxrule=0pt,
    left=4pt, right=4pt,
    top=1pt, bottom=1pt,
    fontupper=\bfseries\footnotesize
  },
  title={Takeaway \stepcounter{takeawaycounter}\thetakeawaycounter}
}

\usepackage{multirow}
\usepackage{makecell}
\title{SAGE: Agentic Framework for Interpretable and Clinically Translatable Computational Pathology Biomarker Discovery}

\author{
Sahar Almahfouz Nasser$^{1}$ \\
Emory University, USA
\And
Juan Francisco Pesantez Borja$^{2}$ \\
Georgia Institute of Technology, USA
\And
Jincheng Liu$^{2}$ \\
Georgia Institute of Technology, USA
\And
Sandeep Manandhar$^{1}$ \\
Emory University, USA
\And
Shikhar Shiromani$^{3}$ \\
NVIDIA, USA
\And
Mohammad Tanvir Hasan$^{4}$ \\
University of Arkansas at Little Rock, USA
\And
Zenghan Wang$^{2}$ \\
Georgia Institute of Technology, USA
\And
Suman Ghosh$^{2}$ \\
Georgia Institute of Technology, USA
\AND
Jinchu Li$^{2}$ \\
Georgia Institute of Technology, USA
\And
Xuejian Xu$^{2}$ \\
Georgia Institute of Technology, USA
\And
Aniket Ramkrishnan Iyer$^{2}$ \\
Georgia Institute of Technology, USA
\And
Naoto Tokuyama$^{1}$ \\
Emory University, USA
\And
Twisha Shah$^{2}$ \\
Georgia Institute of Technology, USA
\And
Tilak Pathak$^{1}$ \\
Emory University, USA
\And
Soundharya Kumaresan$^{1}$ \\
Emory University, USA
\And
Yohei Abe$^{1}$ \\
Emory University, USA
\AND
Himanshu Maurya$^{1}$ \\
Emory University, USA
\And
Anant Madabhushi$^{1}$ \\
Emory University, USA
}
\begin{document}

\maketitle

\begin{abstract}
  Engineered image-based biomarkers offer a clinically interpretable alternative to black-box AI in computational pathology, yet their discovery remains largely intuition-driven, guided by fragmented literature rather than rigorous biological validation. We introduce SAGE (Structured Agentic system for hypothesis Generation and Evaluation), a multi-agent framework that grounds biomarker discovery in biological evidence through three mechanisms: (i) knowledge-graph-anchored hypothesis generation via multi-path ontological reasoning, (ii) a debate-based multi-agent novelty assessment that stress-tests candidate biomarkers against existing literature, and (iii) an end-to-end automated validation pipeline that translates hypotheses directly into executable analyses on multimodal pathology datasets. Together, these components shift biomarker discovery from an intuition-driven, literature-browsing exercise into a structured, traceable reasoning process that clinicians and researchers can inspect, trust, and build upon.
\end{abstract}

\section{Introduction}
\label{sec:introduction}

\begin{figure*}[t]
  \centering
  \includegraphics[width=\textwidth]{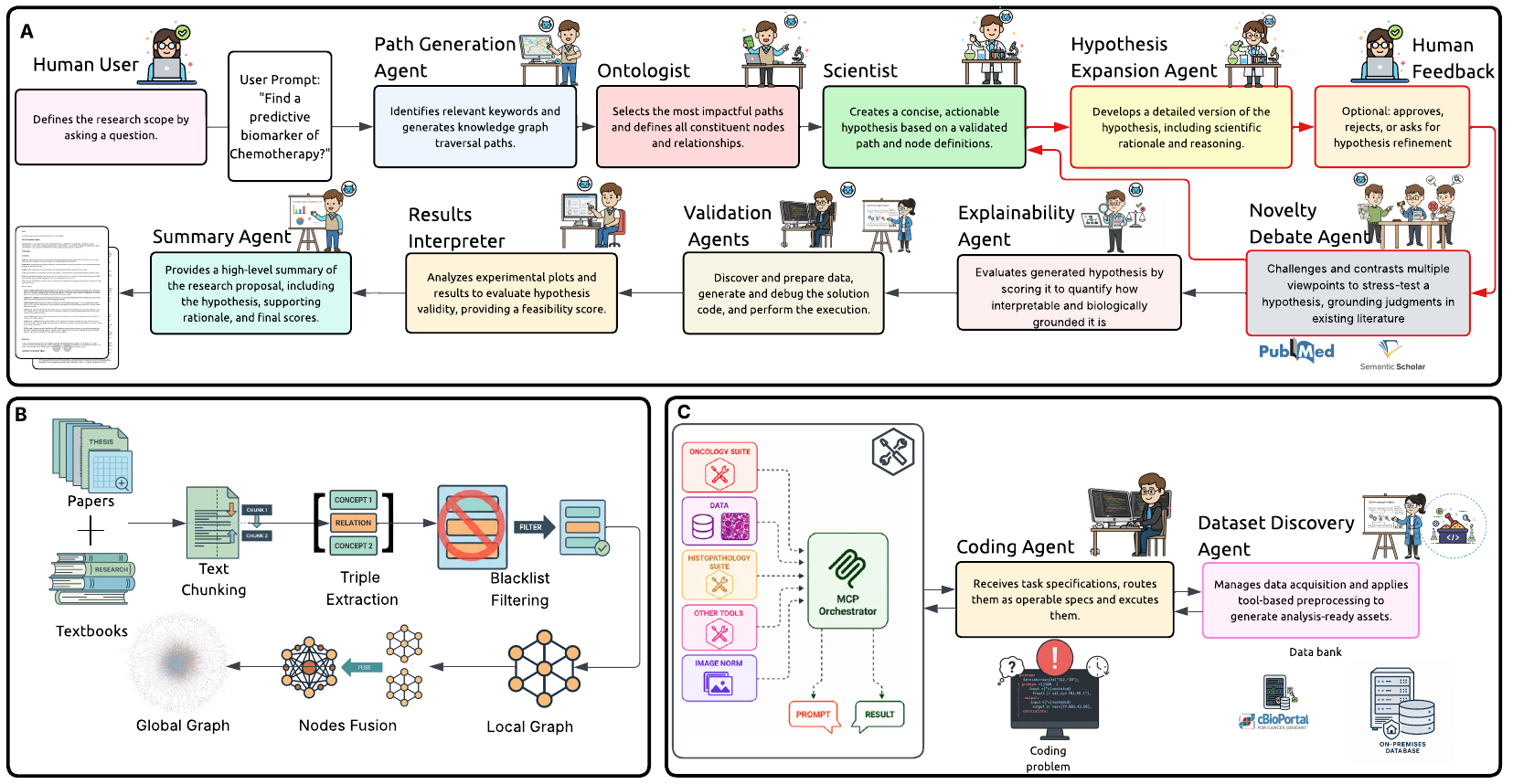}
  \caption{\textbf{Overview of the SAGE framework for computational 
pathology.} The pipeline comprises three stages: 
(A)~\textit{Hypothesis generation and refinement} (top), where 
coordinated agents produce interpretable hypotheses with optional 
human feedback at key decision points; 
(B)~\textit{Knowledge graph construction and reasoning} (bottom 
left), integrating pathology-aware biomedical entities to support 
structured biological reasoning; and 
(C)~\textit{Validation and summarization} (bottom right), which 
executes statistical analyses and generates clinically 
interpretable summaries.}
\label{fig:sage_pipeline}
\end{figure*}

Recent advances in large language models 
(LLMs)~\cite{Luo_2022,singhal2023expertlevelmedicalquestionanswering,
remy2022biolordlearningontologicalrepresentations} have catalyzed 
agentic systems capable of synthesizing scientific literature and 
generating hypotheses. Prominent examples include the Virtual 
Lab~\cite{swanson2025virtual} and AI 
Co-Scientist~\cite{gottweis2025towards}, which demonstrate the 
power of multi-agent reasoning for biomedical discovery. However, 
both systems are designed around wet-lab validation cycles. 
Computational pathology operates under a fundamentally different 
constraint: discoveries must be grounded in measurable features 
spanning imaging, molecular, and clinical domains, and validated 
retrospectively on patient cohort data.

Within computational pathology, a \textit{biomarker} is a 
measurable biological signal that correlates with disease state 
or clinical outcome~\cite{ballman2015biomarker}. Discovering 
reliable biomarkers is central to precision 
medicine~\cite{collins2015new}, yet the field lacks systematic 
tools for doing so. ML-driven systems such as 
SciAgents~\cite{ghafarollahi2024sciagentsautomatingscientificdiscovery} 
demonstrate that coordinated LLM agents can generate interesting 
hypotheses but stop short of biological grounding or empirical 
validation. Clinically motivated frameworks such as 
NOVA~\cite{vaidya2025novaagenticframeworkautomated} and 
WSI-Agents~\cite{lyu2025wsi} target diagnostic objectives but 
rely on black-box models without explicit biological reasoning. 
The result is a field where biomarker discovery remains 
intuition-driven: hypotheses are generated without biological 
justification, and validation is treated as a separate manual 
step.

We introduce \textbf{SAGE} (Structured Agentic system for 
hypothesis Generation and Evaluation), an end-to-end agentic 
framework for biologically grounded biomarker discovery in 
computational pathology. SAGE makes four contributions: 
\textbf{(i)}~an end-to-end discovery pipeline spanning hypothesis 
generation to computational validation on retrospective patient 
cohorts; 
\textbf{(ii)}~multi-path ontological reasoning over a 
domain-specific knowledge graph linking imaging, molecular, and 
clinical concepts; 
\textbf{(iii)}~a debate-based novelty assessment mechanism that 
stress-tests candidate biomarkers against existing literature; 
and \textbf{(iv)}~a biologically grounded validation framework 
that produces interpretable and clinically actionable results.

\section{Related Work}
\label{sec:related_work}

\subsection{Biomarker Discovery in Computational Pathology}
A biomarker is a measurable biological signal, derived from 
tissue morphology, molecular profiles, or their combination, 
that correlates with disease state or clinical 
outcome~\cite{ballman2015biomarker}. Prognostic biomarkers 
stratify patients into high-risk groups (high probability of 
recurrence, relapse, or death) and low-risk groups. Predictive 
biomarkers identify patients likely to respond to a specific 
therapy. Both are central to precision 
oncology~\cite{collins2015new}, where treatment decisions 
depend on patient-specific biological evidence rather than 
population-level statistics.

Recent foundation models such as UNI~\cite{chen2024towards}, 
CONCH~\cite{lu2024visual}, and 
Prov-GigaPath~\cite{xu2024whole} have demonstrated strong 
performance across diagnostic and prognostic tasks. Agentic 
systems such as PathChat~\cite{lu2024multimodal}, 
NOVA~\cite{vaidya2025novaagenticframeworkautomated}, and 
WSI-Agents~\cite{lyu2025wsi} enable interactive slide analysis 
and clinically motivated biomarker discovery. Despite this 
progress, most approaches treat biomarker discovery as a 
pattern recognition problem, producing associations that are 
statistically valid but biologically opaque, difficult to 
interpret, and rarely linked to mechanistic molecular 
hypotheses. SAGE complements these tools by providing a 
structured reasoning process that generates biologically 
grounded hypotheses and validates them computationally on 
multimodal patient data.

\subsection{Agentic Systems for Scientific Discovery}
The AI Co-Scientist~\cite{gottweis2025towards} is the closest 
prior work, employing a multi-agent tournament with 
inference-time scaling to generate and refine biomedical 
hypotheses. SAGE differs in two respects: it is purpose-built 
for computational pathology with domain-specific knowledge 
graphs and tools, and it validates hypotheses entirely in 
silico on retrospective patient cohorts rather than relying 
on wet-lab confirmation. The Virtual 
Lab~\cite{swanson2025virtual} and Agent 
Laboratory~\cite{schmidgall2025agent} demonstrate agentic 
research automation but target wet-lab cycles and general ML 
workflows respectively, making them incomparable to SAGE on 
any shared task. SciAgents~\cite{ghafarollahi2024sciagentsautomatingscientificdiscovery}, 
which directly inspired SAGE, generates hypotheses through 
structured multi-agent reasoning but uses a planner-driven 
architecture with variability across runs and no empirical 
validation. AgentFlow~\cite{li2025intheflowagenticoptimizationeffective} 
similarly uses a dynamic planner requiring on-policy 
reinforcement learning. SAGE instead adopts a fixed serial 
pipeline with well-defined agent roles, requiring no model 
training while ensuring reproducible, biologically grounded 
outputs.

\subsection{Representation of Prior Knowledge and Validation}
Text-based retrieval methods such as 
RAG~\cite{lewis2020retrieval} are effective at capturing 
surface-level evidence but cannot model relationships between 
concepts across domains. Knowledge graphs address this by 
encoding biomedical entities and relationships in structured 
form, enabling multi-hop reasoning. Foundational biomedical 
graphs such as Hetionet~\cite{himmelstein2017systematic} and 
BioKG~\cite{zhang2025comprehensive} support drug-repurposing 
and disease-mechanism discovery, while more recent work such 
as Tree-KG~\cite{niu-etal-2025-tree} and 
GraphMERT~\cite{belova2025graphmertefficientscalabledistillation} 
focuses on scalable KG construction. SAGE adapts these 
principles to computational pathology, building the first 
domain-specific KG connecting imaging, molecular, and clinical 
concepts for hypothesis-driven biomarker discovery. We show 
empirically in Section~\ref{sec:results_kg} that graph-based 
reasoning produces more diverse hypotheses than text-based 
retrieval from a single prompt. On the validation side, ResearchCodeAgent~\cite{gandhi2025researchcodeagentllmmultiagentautomated} integrates LLMs with predefined analysis scripts but its fixed execution logic limits adaptability. SAGE takes a complementary approach: its validation subsystem is equipped with a curated suite of computational pathology tools stress-tested on large multi-institutional cohorts. When a required analysis falls outside this toolset, SAGE's coding agent dynamically generates custom code on demand, ensuring that validation is neither constrained by predefined scripts nor sacrifices the reliability of battle-tested implementations where they exist.



\section{Method}
\label{sec:method}

SAGE operationalizes biomarker discovery as a structured 
sequential pipeline comprising three stages 
(Figure~\ref{fig:sage_pipeline}): knowledge graph construction, 
which grounds all downstream reasoning in curated biological 
evidence; hypothesis generation and evaluation, where coordinated 
agents produce and refine candidate biomarkers; 
and empirical validation, where accepted hypotheses are 
automatically translated into executable analyses on multimodal 
patient cohort data.

\subsection{Knowledge Graph Construction}
\label{subsec:kg_main}

Most agentic systems rely on text-based retrieval for prior 
knowledge, which is constrained to a single query context and 
cannot model relationships across biological domains. SAGE 
instead constructs a domain-specific biomedical knowledge graph 
(KG) encoding entities, relations, and supporting evidence in 
structured form, enabling multi-hop reasoning across imaging, 
molecular, and clinical domains.

\paragraph{Literature Processing and Triple Extraction.}
Candidate papers are screened using expert-curated keyword sets 
and scored on semantic relevance, SapBERT 
similarity~\cite{reimers2019sentencebert}, and publication venue. 
Of all initially retrieved papers, approximately 40\% are 
filtered as insufficiently relevant. Retained PDFs are segmented 
into overlapping 1,500-token chunks, each processed by 
\texttt{GPT-4o-mini} to extract biomedical triples 
$(h, t_h, r_{\text{text}}, r_{\text{norm}}, t, t_t, c, e)$, 
where $h$ and $t$ are head and tail entities, $t_h$ and $t_t$ 
their ontology types, $r_{\text{text}}$ the surface relation, 
$r_{\text{norm}}$ its normalized form, $c \in [0,1]$ a confidence 
score, and $e$ the supporting evidence span. Triples with 
$c \geq 0.5$ are retained and relations normalized via 
ontology-aware rule-based 
mapping~\cite{peng2017cross,zhang2018drug}. See Figure~\ref{fig:kg_pipeline}

\paragraph{Graph Construction and Fusion.}
Local directed graphs $G_d = (V_d, E_d)$ are built per document 
with edges weighted by extraction confidence. Entity redundancy 
across documents is resolved using \texttt{BGE-Large} 
embeddings~\cite{xiao2023bge} at cosine similarity threshold 
$\tau = 0.9$, with redundant edges merged by weight summation. 
Disconnected components are pruned and the resulting global graph 
$\mathcal{G}_{\text{global}}$ serialized in GraphML 
format~\cite{brandes2001graphml}. The final KG integrates 1,650 
sources comprising 41,053 nodes and 56,338 edges. Quality 
assessment confirms 99.0\% factual grounding and 100\% relation 
accuracy, with entity type accuracy improving from 64.6\% to 
82.5\% after ontology refinement 
(Appendix~\ref{app:kg_quality}, Figure~\ref{fig:kg_pipeline}).

\begin{figure}[t]
  \centering
  \includegraphics[width=\textwidth]{Figures/KG.pdf}
  \caption{\textbf{Literature processing pipeline and biomedical 
knowledge graph.} \textit{(Left)} Three-phase pipeline for 
corpus construction: keyword expansion via OpenAI, MeSH, and 
Semantic Scholar APIs; large-scale literature crawling from 
high-impact sources; and automated text extraction and storage.
\textit{(Right)} The resulting biomedical knowledge graph, where 
nodes represent biological entities and edges encode 
relationships mined from literature and medical ontologies. 
Colors indicate entity types (see legend); the zoomed-in 
subgraph highlights dense connectivity among disease-relevant 
entities.}
\label{fig:kg_pipeline}
\end{figure}
\paragraph{Multi-Path Ontological Reasoning.}
\label{sec:path_generation}
Rather than committing to a single KG traversal path, the 
\textbf{Path Generation Agent} identifies multiple candidate 
paths linking entity combinations such as 
\textit{Gene--Pathway--Disease} or 
\textit{Biomarker--Tissue--Outcome}, exposing the breadth of 
biologically plausible connections from a single query. Each 
path $P$ is scored on four metrics balancing plausibility 
against discovery potential, where $\mathbf{v}_Q$ and 
$\mathbf{v}_P$ are query and path embeddings; $\mathbf{e}_i$ 
is the embedding of entity $e_i$ and $\mathbf{c}_{\text{target}}$ 
the target domain centroid; $\deg(e)$ is the degree of entity 
$e$ in the global KG; and $P_{\text{actual}}$, 
$P_{\text{expected}}$ are the entity category distributions 
of the path and full KG respectively:
\begin{align}
S_{\text{logic}} &= \frac{\mathbf{v}_Q \cdot \mathbf{v}_P}
                         {\lVert\mathbf{v}_Q\rVert_2\,\lVert\mathbf{v}_P\rVert_2}
\label{eq:logic}\\[6pt]
S_{\text{rel}} &= 1 - \frac{1}{N}\sum_{i=1}^{N} 
\lVert \mathbf{e}_i - \mathbf{c}_{\text{target}} \rVert_2 
\label{eq:rel}\\[6pt]
S_{\text{nov}} &= \frac{1}{|P|}\sum_{e \in P} 
{-\log P(e)}, \quad 
P(e) = \frac{\deg(e)+\epsilon}{\sum_{v}\deg(v)+\epsilon} 
\label{eq:nov}\\[6pt]
S_{\text{sur}} &= \mathrm{KL}(P_{\text{actual}} 
\,\|\, P_{\text{expected}}) \label{eq:sur}
\end{align}
These are combined into a weighted aggregate:
\begin{equation}
S_{\text{total}} = w_{\text{logic}} S_{\text{logic}} +
w_{\text{rel}} S_{\text{rel}} +
w_{\text{nov}} S_{\text{nov}} +
w_{\text{sur}} S_{\text{sur}}, \quad \sum_i w_i = 1
\label{eq:path_score}
\end{equation}
The weights $w_i$ are user-configurable: higher 
$w_{\text{logic}}$ and $w_{\text{rel}}$ favor plausible 
hypotheses, while higher $w_{\text{nov}}$ and $w_{\text{sur}}$ 
steer SAGE toward bolder discoveries. Here $S_{\text{sur}}$ 
measures distributional surprise relative to background KG 
statistics, capturing paths whose entity-type composition 
deviates meaningfully from the global graph. Full derivations 
are in Appendix~\ref{app:path_scoring}.

\subsection{Hypothesis Generation and Evaluation}
\label{subsec:hypothesis_gen_main}

The hypothesis generation stage coordinates six agents in a fixed 
sequential order, as illustrated in Figure~\ref{fig:sage_pipeline}-A: 
\textbf{Path Generation, Ontologist, Scientist, Hypothesis Expansion, 
Novelty Debate, and Explainability}.
This fixed serial architecture produces consistent, reproducible 
outputs and reduces hallucination risk compared to 
planner-driven 
systems~\cite{ghafarollahi2024sciagentsautomatingscientificdiscovery}, 
while reducing inference cost by up to 75\% through targeted 
context allocation, as shown in Appendix~\ref{app:hypothesis} 
(Table~\ref{tab:pipeline_comparison}).
A human-in-the-loop checkpoint between the Scientist and Hypothesis 
Expansion agents allows domain experts to approve, reject, or refine 
hypotheses before computational resources are committed. Agent roles 
and model assignments are detailed in the Appendix (Table~\ref{tab:workflow_supp}).

The \textbf{Ontologist} selects the most impactful KG paths and 
grounds each in curated biomedical ontologies, mapping graph 
relations to biologically interpretable processes. The 
\textbf{Scientist Agent} then generates concise, dataset-aware 
hypotheses following a structured template that specifies the target 
population, biological and imaging variables, clinical endpoint, and 
expected directionality. The \textbf{Hypothesis Expansion Agent} 
subsequently enriches each approved hypothesis with mechanistic 
rationale and clinical interpretation, producing a fully reasoned 
scientific proposal ready for adversarial evaluation.

\subsubsection{Debate-Based Novelty Assessment}
\label{sec:novelty}

A central contribution of SAGE is the \textbf{Novelty Debate Agent}, 
a multi-critic evaluation framework inspired by adversarial debate 
protocols~\cite{irving2018aisafetydebate} that simulates the peer 
review process through structured argumentation. Three sub-agents 
with distinct epistemic stances participate: a \textbf{Prover} 
(optimistic) that advocates for the hypothesis by identifying unique 
mechanistic claims and gaps in existing literature; a 
\textbf{Verifier} (conservative) that challenges novelty claims 
through real-time retrieval across six scholarly databases (PubMed, 
Semantic Scholar, Europe PMC, bioRxiv, arXiv, and CrossRef); and a 
\textbf{Judge} (balanced) that synthesizes both sides and renders 
a final assessment.

Each critic assigns an initial score on a 1--10 scale. When 
significant disagreement is detected ($\sigma > 1.0$), a multi-round 
Bayesian debate~\cite{zhang2024truthdeceitbayesiandecoding} is 
initiated, updating scores iteratively until convergence 
($\sigma < 0.5$) or a maximum of three rounds. The Verifier can 
flag arguments as \textit{specious} when a novelty claim is directly 
contradicted by retrieved literature, triggering a score penalty 
when upheld by the Judge. The final novelty score is the unweighted 
mean across all three critics.

\subsubsection{Explainability Agent}
\label{sec:explainability}
The \textbf{Explainability Agent} scores each hypothesis on biological and clinical interpretability using an \textit{Explainability Index} (EI) composed of five criteria, each scored 0--2:
\begin{equation}
\mathrm{EI} = \mathrm{MD} + \mathrm{CP} + \mathrm{SBC} + \mathrm{CT} + \mathrm{MT}
\end{equation}
The five components capture complementary dimensions of biomarker credibility: \textbf{Mechanistic Depth (MD)} evaluates grounding in established biological pathways; \textbf{Causal Plausibility (CP)} assesses whether biomarker variation could reasonably influence the disease state or outcome, rather than merely reflecting a downstream correlate; \textbf{Spatial and Biological Coherence (SBC)} checks localization within the relevant tissue architecture; \textbf{Clinical Traceability (CT)} evaluates measurability via pathology-compatible assays; and \textbf{Model Transparency (MT)} assesses whether the biomarker has a clear, reproducible quantitative definition. These dimensions are deliberately non-redundant: for instance, a biomarker mapping to an immune-exhaustion pathway may score high on MD yet low on CP if it appears only as a late consequence of tumor progression rather than a plausible driver of treatment resistance. Conversely, a tumor--immune spatial interaction pattern may score high on CP even when the precise molecular mechanism awaits further validation.

\subsection{Empirical Hypothesis Validation}
\label{subsec:coding_agent}

Accepted hypotheses are forwarded to the validation stage, 
coordinating the \textbf{Validation Agents} subsystem, the 
\textbf{Results Interpreter}, and the \textbf{Summary Agent} 
in sequence.

\subsubsection{Dataset Discovery Agent}

Rather than assuming data availability, the \textbf{Dataset Discovery Agent} 
determines which datasets can support a given hypothesis before any code is 
written. The hypothesis is first parsed into explicit constraints encoding the 
target cohort, predictor, outcome, exposure, and analysis objective. These 
constraints are organized into an \textit{Artifact Requirement Graph} (ARG) 
$\mathcal{A} = (\mathcal{T}, \mathcal{R})$, where $\mathcal{T}$ is a set of 
validation targets (e.g., cohort definition, predictor measurement, outcome 
measurement) and $\mathcal{R}$ is the set of dependencies between them. 
Available files are indexed from public omics repositories, clinical registries, 
and institutional data banks. Each file $f$ is evaluated against the ARG through 
a two-stage deterministic algorithm. In the first stage, each target 
$\tau \in \mathcal{T}$ is decomposed into fine-grained requirements 
$\{r_1, \ldots, r_k\}$ and each file receives a coverage status 
$\phi(f, \tau) \in \{\text{none},\, \text{partial},\, \text{full}\}$, 
reflecting how well the file satisfies the data needs of that target. We further 
define a binary satisfaction indicator:
\begin{equation}
    \sigma(S,\tau) \;:=\; 
\mathbbm{1}\!\left[
    \phi\!\left(\bigcup_{f \in S} f,\,\tau\right) = \text{full}
\right]
    \label{eq:indicator}
\end{equation}
In the second stage, a greedy selection identifies the smallest subset of files 
$\mathcal{F}^* \subseteq \mathcal{F}$ whose combined content fully satisfies 
every ARG target:
\begin{equation}
    \mathcal{F}^* = \arg\min_{S \subseteq \mathcal{F}} |S|
    \quad \text{s.t.} \quad
    \forall\, \tau \in \mathcal{T},\; \sigma(S, \tau) = 1
    \label{eq:greedy}
\end{equation}
where $\bigcup_{f \in S} f$ denotes the union of information contributed by 
all selected files, and $\sigma(S, \tau) = 1$ indicates that target $\tau$ is 
completely satisfied. At each greedy step, the file contributing the highest 
weighted marginal coverage gain over still-unmet targets is added to 
$\mathcal{F}^*$, until all targets reach full coverage or no further gain is 
achievable. Hypotheses for which no suitable $\mathcal{F}^*$ exists are flagged 
for expert review rather than proceeding to code execution.

\subsubsection{Coding Agent and Tool Orchestration}
The \textbf{Coding Agent} operates in three stages: file 
inspection, hypothesis-conditioned code generation, and 
sandboxed execution within containerized environments. 
Failed scripts enter a repair loop using execution feedback, 
retrying up to a fixed budget before escalating to expert 
review (Appendix~\ref{app:attrition}).

Tool selection is mediated by a \textbf{Tool Orchestration 
Subsystem} in which each tool is registered through a 
three-layer specification enabling token-efficient routing, 
planning, and on-demand execution. The orchestrator 
dynamically selects from validated tools spanning survival 
analysis, histopathology, transcriptomics, proteomics, and 
genomic correlation; when no suitable tool exists, the 
Coding Agent synthesizes new Python code within sandbox 
constraints. Full architecture and illustrative examples 
are provided in Appendix~\ref{app:tools}.
\subsubsection{Results Interpreter and Summary Agent}

The \textbf{Results Interpreter} evaluates hypothesis 
validity from Coding Agent outputs based on statistical 
significance, effect size, and biological interpretability. 
The \textbf{Summary Agent} consolidates the hypothesis, 
novelty score, explainability index, and validation results 
into a structured research report.

\section{Experimental Results}
\label{sec:results}

We evaluate SAGE across four dimensions corresponding to its 
core contributions: hypothesis quality and diversity, 
debate-based novelty assessment, automated validation 
performance, and clinical relevance of discovered biomarkers. 
All experiments use bladder cancer as the target disease 
domain, as its biomarker landscape remains comparatively 
underexplored relative to other cancers, making systematic 
discovery particularly valuable.

\subsection{Hypothesis Quality and Diversity}
\label{sec:results_kg}

We compare six configurations to evaluate knowledge graph 
reasoning and multi-path traversal: ChatGPT 5.2 and Gemini 
3.1 Pro as direct prompting baselines; 
AgentFlow~\cite{li2025intheflowagenticoptimizationeffective} 
as a multi-agent baseline (planer-based); SAGE (Text) replacing the KG with 
text retrieval; SAGE (Single-Path) using a single KG path; 
and SAGE (Full) using multi-path ontological reasoning.

Each system generated $N = 10$ hypotheses per prompt across 
46 prompts (questions from four domain experts). For all baselines, SAGE (Text), and SAGE 
(Single-Path), diversity arises from running the same prompt 
10 times independently. For SAGE (Full), diversity arises 
structurally from selecting the top-10 ranked KG paths from 
a single prompt. Diversity is measured using the Vendi 
Score~\cite{friedman2023vendi} computed per prompt ($N=10$, 
max = 10) and averaged across 46 prompts 
(Appendix~\ref{sec:supp_diversity}).

\paragraph{Results.}
Table~\ref{tab:merged_debate_human_results} shows that SAGE 
(Full) achieves the highest automated novelty (7.67) and 
reviewer consensus (7.85). On diversity, SAGE (Full) 
outperforms both SAGE (Single-Path) (5.07 vs 4.11) and SAGE 
(Text) (5.07 vs 4.51), confirming that structured KG 
traversal produces more semantically distinct hypotheses than 
repeated prompting. Although Gemini 3.1 Pro achieves the 
highest Vendi score (6.10), its lower automated novelty 
(6.90) and expert novelty (4.53) suggest this reflects 
stochastic generation variability rather than biologically 
structured exploration, see the limitations of Vendi score in Appendix~\ref{sec:supp_diversity}.

Expert evaluation by four blinded domain specialists 
confirms that SAGE (Full) achieves the highest novelty score 
(7.10). A tradeoff emerged: novel hypotheses scored lower on 
feasibility because experts were less familiar with the 
proposed associations, while ChatGPT and Gemini generated 
hypotheses resembling reformulations of the original 
questions, appearing more immediately feasible. This tradeoff 
is consistent with SAGE's goal of surfacing underexplored 
rather than established associations, please refer to Appendix~\ref{app:expert_eval} for more details on the expert evaluation protocol.

\begin{table*}[t]
\centering
\caption{Hypothesis quality evaluation across six 
configurations ($N = 10$ hypotheses per prompt, 46 prompts). 
Debate-agent scores: novelty, confidence, and consensus 
(1--10). Vendi Score: per-prompt diversity averaged across 
46 prompts (max = 10). Human-expert scores: novelty and 
feasibility from four blinded domain specialists. 
\textbf{Bold} = best per metric. 
$^*$ = significantly different from SAGE (Full) 
based on non-overlapping 95\% confidence intervals.}
\label{tab:merged_debate_human_results}
\scriptsize
\setlength{\tabcolsep}{2.8pt}
\renewcommand{\arraystretch}{1.1}
\resizebox{\textwidth}{!}{%
\begin{tabular}{lcccccc}
\toprule
\multirow{2}{*}{\textbf{System}}
& \multicolumn{4}{c}{\textbf{Debate Agent Scores}}
& \multicolumn{2}{c}{\textbf{Human Expert Scores}} \\
\cmidrule(lr){2-5} \cmidrule(lr){6-7}
& \textbf{Novelty}
& \textbf{Confidence}
& \textbf{Consensus}
& \textbf{Vendi}
& \textbf{Novelty}
& \textbf{Feasibility} \\
\midrule
ChatGPT 5.2
& 6.91$^*$ [6.75, 7.07]
& \textbf{6.70$^*$} [6.49, 6.90]
& 7.06
& 5.00
& 6.74 [5.95, 7.53]
& \textbf{10.00$^*$} [10.00, 10.00] \\
Gemini 3.1 Pro
& 6.90$^*$ [6.72, 7.08]
& 6.30 [6.08, 6.52]
& 7.23
& \textbf{6.10}
& 4.53$^*$ [4.08, 4.98]
& 8.56$^*$ [8.18, 8.94] \\
AgentFlow
& 6.82$^*$ [6.63, 7.01]
& \textbf{6.76$^*$} [6.52, 6.99]
& 6.76
& 5.22
& 4.83$^*$ [4.24, 5.42]
& 8.27$^*$ [7.73, 8.81] \\
SAGE (Text)
& 7.00$^*$ [6.87, 7.13]
& 6.18 [6.04, 6.32]
& 7.21
& 4.51
& 5.41$^*$ [4.89, 5.93]
& 7.26 [6.79, 7.73] \\
SAGE (Single-Path)
& 7.33 [7.16, 7.50]
& 6.42$^*$ [6.25, 6.60]
& 7.18
& 4.11
& 6.36 [5.64, 7.08]
& 7.86 [7.41, 8.31] \\
SAGE (Full)
& \textbf{7.67} [7.50, 7.84]
& 6.00 [5.91, 6.09]
& \textbf{7.85}
& 5.07
& \textbf{7.10} [6.49, 7.71]
& 7.03 [6.54, 7.52] \\
\bottomrule
\end{tabular}
}
\end{table*}

\begin{takeaway}
SAGE (Full) achieved the highest novelty and improved diversity compared with the single-path and text-based SAGE variants, highlighting the value of multi-path knowledge graph reasoning.
\end{takeaway}

\subsection{Ablation Study: Debate-Based Novelty Assessment}
\label{sec:results_debate}

We evaluate whether adversarial deliberation, defined as 
iterative evidence-driven score revision triggered by critic 
disagreement, improves novelty assessment calibration beyond 
multi-critic averaging. All experiments use time-travel 
backtesting on 150 historical proposals spanning five novelty 
tiers to prevent temporal 
leakage~\cite{ye2026prooftimebenchmarkevaluating}. Full 
dataset composition and deliberation statistics are provided 
in Appendix~\ref{sec:supp_novelty}.

\paragraph{Experimental design.}
We compare four configurations isolating the effect of 
deliberation: (i)~\emph{Single-Critic}; (ii)~\emph{Two-Critic} 
ensemble (Prover and Verifier) without interaction; 
(iii)~\emph{Multi-Critic (No Debate)}, where three critics 
score independently and are averaged; and 
(iv)~\emph{Multi-Critic (Full)}, our proposed design, where 
critics enter iterative Bayesian debate when initial 
disagreement exceeds $\sigma > 1.0$. The sole difference 
between (iii) and (iv) is disagreement-triggered deliberation, 
isolating the contribution of our debate mechanism.

\paragraph{Results.}
Table~\ref{tab:separability} reports overall accuracy and 
A--E separability, defined as the mean score difference 
between breakthrough (A) and specious (E) proposals. 
Multi-Critic (Full) achieves the largest marginal gain: 
a 1.76-point increase in A--E separability over the no-debate 
condition ($^*p < 0.05$), and a 35.3\% absolute accuracy 
improvement over single-critic scoring. Deliberation is 
triggered most frequently for extreme categories (A: 76.7\%, 
E: 83.3\%), where it resolves high initial disagreement 
through structured evidence exchange. Error analysis and computational cost 
breakdown are provided in Appendix~\ref{sec:supp_novelty}.

\begin{table}[t]
\centering
\caption{Effect of deliberation on novelty separability 
across 150 proposals. A--E Gap denotes the mean score 
difference between breakthrough (A) and specious (E) 
proposals. $^*$Statistically significant improvement 
over the preceding configuration ($p < 0.05$).}
\label{tab:separability}
\small
\begin{tabular}{lcc}
\toprule
\textbf{Configuration} & \textbf{Accuracy} & 
\textbf{A--E Gap} \\
\midrule
Single-Critic                  & 56.0\%               & 1.12 \\
Two-Critic (Verifier + Prover) & 75.3\%$^*$           & 4.18$^*$ \\
Multi-Critic (No Debate)       & 81.3\%$^*$           & 5.02$^*$ \\
\textbf{Multi-Critic (Full)}   & \textbf{91.3\%}$^*$ & 
\textbf{6.78}$^*$ \\
\bottomrule
\end{tabular}
\end{table}

\begin{takeaway}
Debate across three critics improves novelty assessment accuracy by 35.3\% over a single agent.
\end{takeaway}

\subsection{Ablation Study: Validation Pipeline}
\label{sec:results_validation}

We evaluate the importance of dedicated dataset discovery by 
benchmarking the Coding Agent on 
KramaBench~\cite{lai2026kramabench}, a public data-science 
benchmark spanning diverse tabular reasoning tasks. We compare 
two conditions that bound the impact of data retrieval quality: 
\textbf{(i) Oracle Discovery} ($N=94$ tasks), where the 
task-relevant files are pre-attached to the Coding Agent, 
simulating a perfect file-selection oracle and establishing 
an upper bound on achievable performance; and \textbf{(ii) 
Unguided Discovery} ($N=106$ tasks), where the Coding Agent 
autonomously searches a raw dataset root without any structured 
retrieval, representing the lower bound when no dedicated 
discovery mechanism is available.

Table~\ref{tab:coding_ablation} summarizes performance across 
both conditions and four model backends. Under oracle 
conditions, execution success rates range from 61.7\% to 
78.7\%, reflecting the Coding Agent's capability when 
supplied with correct data. Under unguided discovery, 
execution rates drop sharply to 29.2\%--39.6\%, with the 
dominant failure mode being failure to locate a valid file 
set rather than failures in code generation or execution. 
This gap directly demonstrates that data retrieval, not 
code generation, is the primary bottleneck in end-to-end 
hypothesis validation. These findings motivated the design 
of the Dataset Discovery Agent in SAGE, which replaces 
unguided file search with a deterministic ARG-based 
retrieval algorithm that formally matches hypothesis 
requirements to available patient data. Note that 
Table~\ref{tab:coding_ablation} does not directly measure 
the performance of SAGE's Dataset Discovery Agent, but 
rather quantifies the performance headroom that a 
structured discovery mechanism can unlock. Full attrition 
analysis and per-model breakdown are provided in 
Appendix~\ref{app:attrition}.

\begin{table}[t]
\centering
\caption{Coding Agent performance on KramaBench under 
oracle file selection (upper bound) and unguided autonomous 
search (lower bound) across four model backends. 
\textit{Exec.\%}: fraction of tasks where the generated 
script ran successfully. \textit{Exact\%}: fraction of 
tasks with a correct answer. The gap between conditions 
quantifies the performance headroom unlocked by dedicated 
dataset discovery.}
\label{tab:coding_ablation}
\small
\setlength{\tabcolsep}{4pt}
\renewcommand{\arraystretch}{1.1}
\begin{tabular}{lcccc|lcccc}
\toprule
\multicolumn{5}{c|}{\textbf{Oracle Discovery} ($N=94$)} & 
\multicolumn{5}{c}{\textbf{Unguided Discovery} ($N=106$)} \\
\midrule
\textbf{Model} & \textbf{Score} & \textbf{Exact\%} & 
\textbf{Exec.\%} & \textbf{Failure} &
\textbf{Model} & \textbf{Score} & \textbf{Exact\%} & 
\textbf{Exec.\%} & \textbf{Failure} \\
\midrule
codex     & 42.18 & 31.9\% & 78.7\% & Script &
codex     & 18.28 & 12.3\% & 37.7\% & No files \\
5.4-03-05 & 40.79 & 28.7\% & 73.4\% & Script &
5.4-03-05 & 15.76 &  9.4\% & 29.2\% & No files \\
5.4-mini  & 32.04 & 25.5\% & 63.8\% & Script &
5.4-mini  & 12.89 &  9.4\% & 39.6\% & Exec. \\
5.4-nano  & 29.90 & 20.2\% & 61.7\% & Script &
5.4-nano  & 12.96 &  9.4\% & 37.7\% & No files \\
\bottomrule
\end{tabular}
\end{table}

\begin{takeaway}
Poor data retrieval, not code generation, is the primary bottleneck in automated hypothesis validation.
\end{takeaway}

\subsection{Case Study: End-to-End Biomarker Discovery}
\label{sec:results_case_study}

Starting from a single research question on prognostic 
signatures in bladder cancer, SAGE autonomously traversed 
its knowledge graph, generated and stress-tested a hypothesis, 
and executed its validation on 412 TCGA-BLCA patients without 
manual intervention.

\textbf{Discovery.} SAGE identified a joint prognostic 
interaction between FABP5 expression and the abundance of 
tertiary lymphoid structures (TLS), immune aggregates 
quantified from whole-slide images, a multimodal biomarker 
spanning molecular and imaging domains absent from prior 
literature.

\begin{figure}[t]
    \centering
    \includegraphics[width=\linewidth]{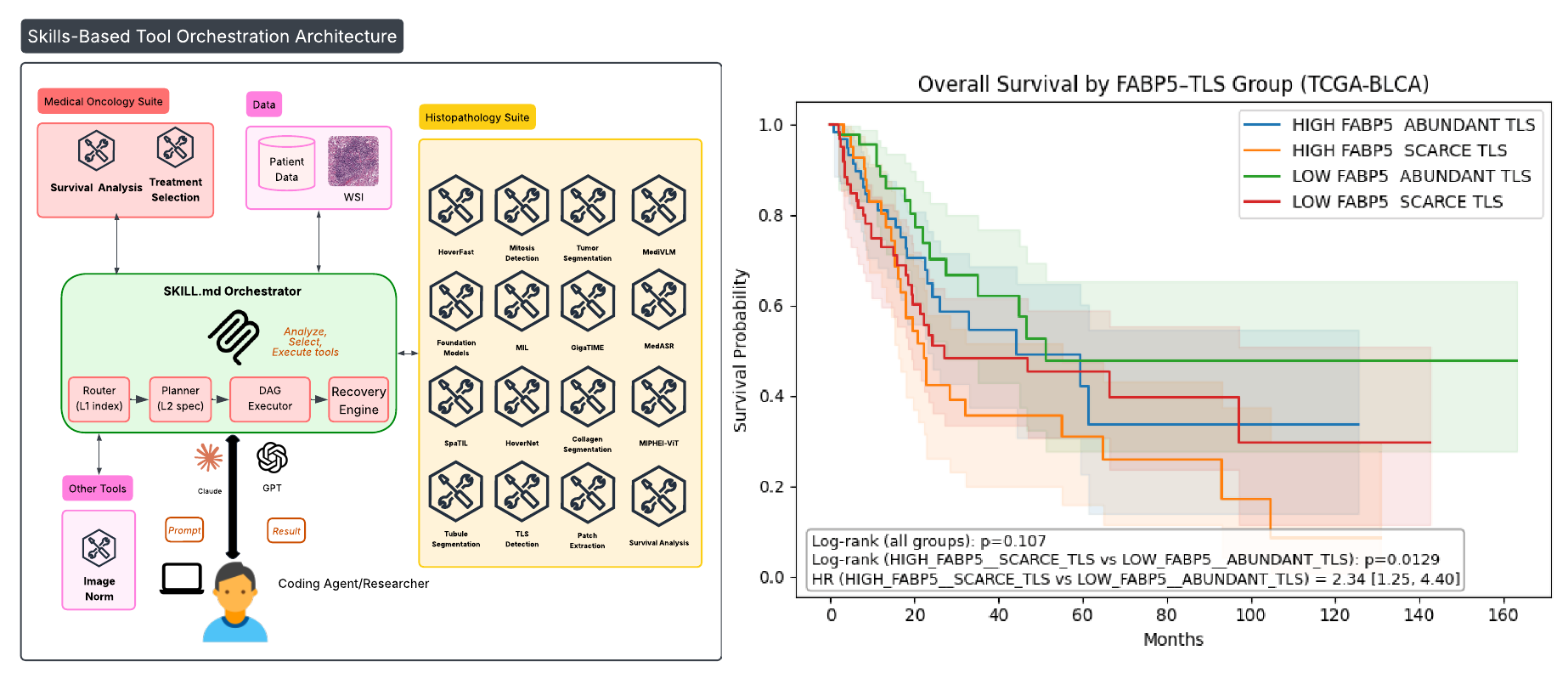}
   \caption{\textbf{Tool orchestration and end-to-end validation.} 
\textit{(Left)} The SKILL.md orchestrator routes, plans, and 
executes tools across oncology and histopathology suites. 
\textit{(Right)} Overall survival by joint FABP5 and TLS 
stratification in TCGA-BLCA (HR = 2.34 [1.25, 4.40], $p = 0.013$).}

    \label{fig:fabp5_tls_km}
\end{figure}

\textbf{Validation.} Patients with high FABP5 and scarce 
TLS exhibited significantly worse overall survival compared 
to those with low FABP5 and abundant TLS 
(Figure~\ref{fig:fabp5_tls_km}), confirmed by Cox 
proportional hazards modeling. Additional validated 
hypotheses generated by SAGE are provided in 
Appendix~\ref{app:results}.

\begin{takeaway}
SAGE autonomously discovered and validated a novel multimodal biomarker, absent from prior literature.
\end{takeaway}




\section{Conclusion}
\label{sec:conclusion}

SAGE treats biomarker discovery as a structured reasoning 
problem, unifying knowledge-graph construction, multi-path 
hypothesis generation, adversarial novelty debate, and 
automated in silico validation in a single pipeline. 
Multi-path graph-based reasoning produced significantly 
more diverse hypotheses than text-based retrieval, 
three-critic debate improved novelty assessment by 35.3\% 
over single-agent scoring, and a dedicated dataset 
discovery agent improved validation success by 40\%. 
Applied to bladder cancer, SAGE autonomously discovered 
and validated novel multimodal biomarkers 
absent from prior literature.

Current limitations include the restriction of the 
knowledge graph to text-based sources, the absence of 
direct benchmarking against systems such as AI 
Co-Scientist~\cite{gottweis2025towards} due to unavailable 
code, and the use of a general-purpose benchmark as a 
proxy for validation pipeline evaluation. Future work will 
extend the knowledge graph to genomic, proteomic, and 
spatial transcriptomic data across multiple cancer types, 
and quantify validation failure modes on real computational 
pathology tasks. Furthermore, when adapting SAGE to new scientific domains, 
future work will introduce an automated model discovery 
module that searches online repositories such as GitHub 
for domain-validated computational models, dynamically 
constructing a domain-specific tool registry for the 
coding agent rather than relying on a manually curated 
set of tools.

The design principles of SAGE are domain-agnostic and 
transferable to any scientific domain requiring 
interpretable, empirically grounded discovery, with the 
automated tool registry construction serving as the key 
bridge for seamless domain adaptation.

\bibliographystyle{plainnat}
\bibliography{example_paper}  

\newpage
\appendix

\section{Knowledge Graph Construction}
\label{app:kg}

\subsection{Literature Processing Details}
\label{app:kg_extraction}

Candidate papers are retrieved using domain-specific keyword 
queries covering computational pathology, disease biomarkers, 
spatial analysis, and whole-slide imaging. Each paper is scored 
using a composite metric combining keyword frequency, semantic 
proximity via SapBERT similarity, and journal impact factor. 
Thresholds are calibrated empirically to balance recall and 
precision, resulting in approximately 40\% of retrieved papers 
being excluded as insufficiently relevant.

PDF text extraction uses a three-tier fallback strategy 
employing \texttt{pdfplumber}, \texttt{PyMuPDF}, and 
\texttt{pypdf} to accommodate heterogeneous document formats. 
Extracted text is normalized and segmented into overlapping 
chunks of approximately 1,500 tokens to preserve local semantic 
context for relation extraction.

Triple extraction uses \texttt{GPT-4o-mini} with structured 
prompts enforcing entity typing, relation normalization, and 
evidence attribution. The ontology schema includes fine-grained 
entity types such as \textit{Gene}, \textit{Pathway}, 
\textit{Disease}, \textit{ClinicalEndpoint}, 
\textit{TissueRegion}, \textit{StainingMethod}, and 
\textit{Algorithm}. Disambiguation rules resolve patient 
cohorts, clinical outcomes, and gene identifiers.

\subsection{Graph Construction Details}
\label{app:kg_construction}

For each document $d$, a local graph $G_d = (V_d, E_d)$ is 
constructed where nodes correspond to unique entities and 
edges aggregate all triples linking entity pairs, with weights 
defined as the maximum confidence among supporting triples. 
Local graphs are fused into a global KG using 
\texttt{BGE-Large} embeddings~\cite{xiao2023bge} with cosine 
similarity threshold $\tau = 0.9$. Redundant edges are merged 
by summing weights and disconnected components with fewer than 
three nodes are pruned. Graphs are serialized in GraphML 
format~\cite{brandes2001graphml,hagberg2008networkx}.

\begin{figure}[h]
\centering
\includegraphics[width=\columnwidth]{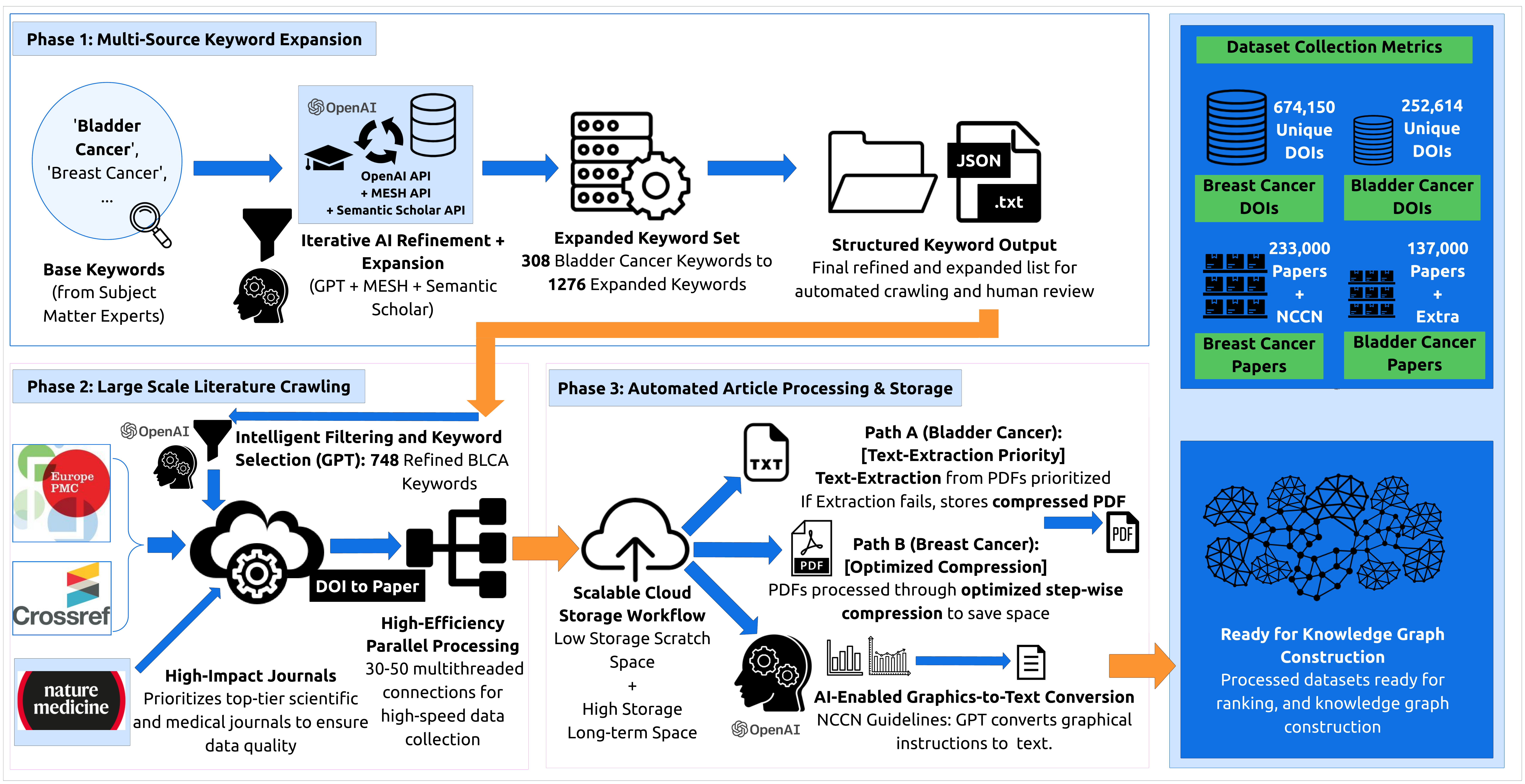}
\caption{Three-phase literature processing pipeline for KG 
construction: keyword expansion via OpenAI, MeSH, and Semantic 
Scholar APIs; large-scale crawling from high-impact sources; 
and automated text extraction and storage.}
\label{fig:data_prep_visualization}
\end{figure}

\subsection{Knowledge Graph Quality Assessment}
\label{app:kg_quality}

Quality is assessed via stratified random sampling of 200 
triples across relation types, evaluated on three dimensions: 
factual grounding with respect to the cited evidence span, 
correctness of entity names and types, and accuracy of the 
extracted relation. Results are summarized in 
Table~\ref{tab:kg_quality_supp}.

\begin{table}[h]
\centering
\caption{Knowledge graph quality assessment on a stratified 
random sample of 200 extracted triples.}
\label{tab:kg_quality_supp}
\small
\begin{tabular}{lcc}
\toprule
\textbf{Metric} & \textbf{Initial} & \textbf{Refined} \\
\midrule
Factual grounding          & 99.0\% & 99.0\% \\
Relation accuracy          & 100.0\% & 100.0\% \\
Entity type accuracy (strict) & 64.6\% & 82.5\% \\
Entity type accuracy (lenient) & 99.0\% & 92.5\% \\
Hallucination rate         & 0.5\% & $<$1.0\% \\
\bottomrule
\end{tabular}
\end{table}

Remaining challenges include entity boundary delineation for 
complex noun phrases and disambiguation of clinical endpoints, 
motivating future integration of ontology-constrained 
classifiers and external resources such as 
UMLS~\cite{bodenreider2004unified}.

\subsection{Path Scoring Metric Derivations}
\label{app:path_scoring}

This section provides full derivations of the four path-level 
metrics introduced in Section~\ref{sec:path_generation}.

\subsubsection{Logic Score}
Each path $P$ is converted into a natural-language sentence by 
concatenating its entities and relations in traversal order. 
Both the query $Q$ and path description are embedded into a 
shared semantic space. The Logic score is the cosine similarity 
between query embedding $\mathbf{v}_Q$ and path embedding 
$\mathbf{v}_P$:
\begin{equation}
S_{\text{logic}} = \mathrm{CosSim}(\mathbf{v}_Q, \mathbf{v}_P)
= \frac{\mathbf{v}_Q \cdot \mathbf{v}_P}
{\lVert\mathbf{v}_Q\rVert \lVert\mathbf{v}_P\rVert}
\end{equation}
Higher values indicate stronger semantic alignment; lower 
values indicate conceptual gaps or implausible transitions.

\subsubsection{Relevance Score}
Each entity $e_i$ in path $P = \{e_1, \ldots, e_N\}$ is 
represented by normalized embedding $\mathbf{e}_i$, and the 
target domain by centroid $\mathbf{c}_{\text{target}}$. 
Relevance is the average proximity of path entities to the 
domain centroid:
\begin{equation}
S_{\text{rel}} = 1 - \frac{1}{N} \sum_{i=1}^{N} 
\lVert \mathbf{e}_i - \mathbf{c}_{\text{target}} \rVert_2
\end{equation}
The score is linearly rescaled and clamped to $[0,1]$ for 
comparability across paths.

\subsubsection{Novelty Score}
The Novelty score favors paths containing rare, underexplored 
entities. An empirical occurrence probability is defined for 
each entity based on its degree in the global KG:
\begin{equation}
P(e) = \frac{\deg(e) + \epsilon}{\sum_{v \in V} \deg(v) + 
\epsilon}
\end{equation}
where $\epsilon$ is a small smoothing constant. The Novelty 
score is the average information content of path entities:
\begin{equation}
S_{\text{nov}} = \frac{1}{|P|} \sum_{e \in P} {-\log P(e)}
\end{equation}
Scores are normalized using min-max scaling across candidate 
paths.

\subsubsection{Surprise Score}
The Surprise score captures atypical cross-domain entity 
combinations. Let $P_{\text{actual}}(c)$ denote the entity 
category distribution within path $P$ and 
$P_{\text{expected}}(c)$ the global KG category distribution. 
Surprise is the KL divergence between these distributions:
\begin{equation}
S_{\text{sur}} = \mathrm{KL}(P_{\text{actual}} \,\|\, 
P_{\text{expected}}) = \sum_{c} P_{\text{actual}}(c) 
\log\frac{P_{\text{actual}}(c)}{P_{\text{expected}}(c)}
\end{equation}
Divergence values are normalized to $[0,1]$ across all 
candidate paths. Higher values indicate paths that combine 
entity categories in unexpected ways.

\subsubsection{Weighted Aggregation}
The four normalized metrics are combined as:
\begin{equation}
S_{\text{total}} = w_{\text{logic}} S_{\text{logic}} +
w_{\text{rel}} S_{\text{rel}} +
w_{\text{nov}} S_{\text{nov}} +
w_{\text{sur}} S_{\text{sur}}, \quad \sum_i w_i = 1
\end{equation}
In our experiments, higher weights are assigned to Novelty 
and Surprise to emphasize discovery-oriented criteria. 
Candidate paths are ranked in descending order and the 
top-ranked paths forwarded to the Ontologist agent.

\section{Hypothesis Generation Pipeline}
\label{app:hypothesis}

\subsection{Agent Descriptions and Model Assignments}
\label{app:agents}

Table~\ref{tab:workflow_supp} details the role, context source, 
model assignment, and model-selection rationale for each agent 
in the hypothesis generation pipeline. The fixed serial 
architecture assigns each agent only task-relevant context from 
its immediate predecessor, avoiding the token overhead and 
contextual noise of shared-memory approaches.

\begin{table*}[ht]
\centering
\caption{Complete agent workflow. Each agent is defined by its 
context source in the graph-based pipeline (GP), model 
assignment, role, and model-selection rationale.}
\label{tab:workflow_supp}
\small
\setlength{\tabcolsep}{3.5pt}
\begin{tabular}{lp{2.3cm}p{1.6cm}p{4.8cm}p{4.2cm}}
\toprule
\textbf{Agent} & \textbf{Context Source} & \textbf{Model} & 
\textbf{Role} & \textbf{Rationale} \\
\midrule
Path Generation & Query only & GPT-5 Nano & 
Generate paths between KG nodes via graph traversal. & 
Cheapest model; requires only tool calling with minimal 
reasoning. \\[3pt]
Ontologist & Path Generation & GPT-5 Mini & 
Define biological relationships and technical terms in 
generated paths. & 
Balanced model; domain knowledge required but not maximal 
reasoning. \\[3pt]
Scientist & Ontologist & GPT-5 (high reasoning) & 
Formulate concise, dataset-aware, empirically testable 
hypotheses. & 
Best available model; critical reasoning task determining 
overall hypothesis quality. \\[3pt]
Hypothesis Expansion & Scientist, Ontologist & GPT-5 Mini & 
Enrich hypothesis with scientific rationale, background, 
and clinical implications. & 
Balanced model; competence needed but conciseness valued 
over maximal reasoning. \\[3pt]
Novelty Debate & Hypothesis Expansion & GPT-5 Mini & 
Stress-test hypothesis novelty through adversarial 
three-critic debate grounded in real-time literature 
retrieval. & 
Balanced model; requires literature search and comparative 
analysis. \\[3pt]
Explainability & Hypothesis Expansion & GPT-5 Mini & 
Score hypothesis on five interpretability criteria to 
compute the Explainability Index (EI). & 
Balanced model; structured scoring task requiring domain 
awareness. \\[3pt]
Dataset Discovery & Hypothesis Expansion & 
Deterministic & 
Parse hypothesis into explicit constraints, construct the 
Artifact Requirement Graph (ARG), and identify the minimal 
file set satisfying all validation requirements via greedy 
selection. & 
Rule-based deterministic algorithm; no LLM required as 
retrieval is governed by explicit constraint matching and 
coverage scoring. \\[3pt]
Coding & Dataset Discovery, Hypothesis Expansion & 
GPT-5 (high reasoning) & 
Synthesize and execute hypothesis-conditioned analysis 
code on selected patient cohort data within a sandboxed 
environment. & 
Best available model; code generation and debugging 
require strong reasoning and adaptability. \\[3pt]
Results Interpreter & Coding, Hypothesis Expansion, 
Dataset Discovery & 
GPT-5 (high reasoning + visual) & 
Interpret statistical outputs, assess validation outcomes, 
and assign a feasibility score. & 
Multimodal high-reasoning model; requires quantitative 
and visual analysis. \\[3pt]
Summary & Results Interpreter, Hypothesis Expansion, 
Novelty Debate & GPT-5 Mini & 
Synthesize hypothesis, rationale, novelty score, EI, and 
feasibility score into a structured report. & 
Balanced model; synthesis task prioritizing clarity and 
structured reporting. \\
\bottomrule
\end{tabular}
\end{table*}

\subsection{Pipeline Architecture and Context Allocation}
\label{sec:pipeline_context}

SAGE uses a specific-memory graph-based pipeline (GP) in which 
each agent receives only task-relevant context from its immediate 
predecessor. This contrasts with a shared-memory chat-based 
pipeline (CP) in which every agent receives the full conversation 
history. Table~\ref{tab:pipeline_comparison} shows that GP 
reduces prompt tokens by 57.5\% and completion tokens by 14.9\%, 
cutting inference cost by 25--75\% while maintaining comparable 
hypothesis novelty (7/10 vs.\ 8/10). These results confirm that 
targeted context allocation preserves discovery quality while 
enabling predictable, scalable execution.

\begin{table}[h]
\centering
\caption{Comparison of graph-based (GP) and chat-based (CP) 
pipelines. GP reduces token usage and cost while maintaining 
comparable hypothesis quality.}
\label{tab:pipeline_comparison}
\small
\setlength{\tabcolsep}{4pt}
\begin{tabular}{lccc}
\toprule
\textbf{Metric} & \textbf{GP} & \textbf{CP} & 
\textbf{Change} \\
\midrule
Prompt tokens (SD)     & 9,002 $\pm$ 2,367  & 
21,210 $\pm$ 3,146 & $-$57.5\% \\
Completion tokens (SD) & 10,011 $\pm$ 2,909 & 
11,765 $\pm$ 817   & $-$14.9\% \\
Runtime sec. (SD)      & 222 $\pm$ 73       & 
227 $\pm$ 29       & $-$2.2\%  \\
Novelty score          & 7/10               & 
8/10               & $-$12.5\% \\
Relative cost          & 1$\times$          & 
1.3--4$\times$     & 25--75\%$\downarrow$ \\
\bottomrule
\end{tabular}
\end{table}

The rationale for excluding information at each stage is that 
upstream content is already distilled into the predecessor 
agent's output, making re-exposure redundant and increasing 
the risk of context dilution, where long prompts degrade LLM 
focus and increase hallucination. Table~\ref{tab:context_allocation} 
details the context provided and excluded for each agent.

\begin{table*}[h]
\centering
\caption{Context allocation per agent in the specific-memory 
pipeline. Each agent receives only the information essential 
to its function.}
\label{tab:context_allocation}
\small
\setlength{\tabcolsep}{3pt}
\begin{tabular}{p{2.8cm}p{4.8cm}p{4.8cm}p{3.5cm}}
\toprule
\textbf{Agent} & \textbf{Context Provided} & 
\textbf{Context Excluded} & \textbf{Rationale} \\
\midrule
Path Generation & User query & None (pipeline start) & 
Needs only the query to traverse the KG. \\[2pt]
Ontologist & All paths from Path Generation & 
User query & 
Paths already encode query intent. \\[2pt]
Scientist & Ontologist output & 
Path Generation output & 
Ontologist already refines path information. \\[2pt]
Hypothesis Expansion & Scientist hypothesis & 
Ontologist, Path Generation outputs & 
Scientist hypothesis synthesizes all upstream content. \\[2pt]
Novelty Debate & Expanded hypothesis & 
All upstream outputs & 
Needs only final hypothesis for literature comparison. \\[2pt]
Explainability & Expanded hypothesis & 
All upstream outputs & 
Scores hypothesis independently of generation history. \\[2pt]
Dataset Discovery & Expanded hypothesis & 
All upstream outputs & 
Deterministic algorithm; requires only hypothesis 
constraints to construct ARG and retrieve matching files. \\[2pt]
Coding & Expanded hypothesis, selected file set 
from Dataset Discovery & 
All upstream conceptual outputs & 
Requires hypothesis variables and identified data files 
to synthesize and execute analysis code. \\[2pt]
Results Interpreter & Expanded hypothesis, coding 
outputs, and selected file set & 
Upstream conceptual outputs & 
Requires hypothesis, execution results, and data 
context for statistical interpretation. \\[2pt]
Summary & Results Interpreter, Hypothesis Expansion, 
Novelty Debate & 
Coding outputs, upstream conceptual outputs & 
Needs validation results, final hypothesis, and novelty 
score only. \\
\bottomrule
\end{tabular}
\end{table*}


\subsection{Hypothesis Diversity Evaluation: Graph-Based 
vs. Text-Based Retrieval}
\label{sec:supp_diversity}

\paragraph{Diversity Metric: Vendi Score.}
We adopt the Vendi Score~\cite{friedman2023vendi} as our 
primary diversity metric. Given $N$ hypotheses embedded 
as vectors $\{x_1, \ldots, x_N\}$, the Vendi Score is 
defined as:
\begin{equation}
\text{VS}(X) = \exp\!\left(-\sum_{i=1}^{N} \lambda_i 
\log \lambda_i\right)
\end{equation}
where $\{\lambda_i\}$ are the eigenvalues of the normalized 
cosine kernel matrix $\tilde{K} = K/N$, with $K_{ij} = 
k(x_i, x_j)$ the pairwise cosine similarity. The Vendi 
Score measures the \textit{effective number of independent 
semantic modes} in the hypothesis set: a value of 1 
indicates all hypotheses are identical, while a value of 
$N$ indicates they are pairwise orthogonal. Hypotheses 
forming $k$ tight semantic clusters yield VS $\approx k$.

We prefer the Vendi Score over alternatives such as 
pairwise cosine distance (PCD), Self-BLEU, or LLM judges 
for three reasons: (i) it measures the number of distinct 
semantic centers rather than average spread, capturing 
cluster structure that PCD cannot; (ii) it is insensitive 
to surface lexical variation that inflates Self-BLEU-based 
metrics; and (iii) it is fully deterministic given fixed 
hypotheses and embedding model, introducing no LLM sampling 
variance. Chen et al.~\cite{chen2026diversity} report that 
the Vendi Score achieves 87\% agreement with expert human 
diversity judgments across 500 pairwise comparisons, the 
highest among four evaluated metrics. In our experiments, 
the Vendi Score is computed per prompt on $N = 10$ 
hypotheses using \texttt{text-embedding-3-small} embeddings 
with L2 normalization, and averaged across all 46 prompts. 
Eigenvalues are computed via \texttt{numpy.linalg.eigvalsh} 
with a numerical floor of $10^{-12}$.

\paragraph{Limitations.}
The Vendi Score operates on full hypothesis text rather 
than extracted biological entities, meaning mechanistically 
similar hypotheses phrased differently may appear more 
diverse than they are. Both conditions share the same 
embedding model, ensuring the comparison is controlled. 
Absolute Vendi values are not directly comparable across 
embedding models, though relative orderings are robust to 
embedding choice~\cite{chen2026diversity}.

\subsection{Novelty Debate: Extended Evaluation}
\label{sec:supp_novelty}

\subsubsection{Evaluation Dataset Composition}

We constructed a ground-truth dataset of 150 scientific 
proposals spanning five novelty tiers using time-travel 
backtesting, evaluating each proposal only against literature 
available prior to its publication or retraction date. 
Category A (\emph{Breakthrough}) includes Nobel-winning 
discoveries and paradigm-shifting methods such as AlphaFold2, 
CRISPR--Cas9, Transformer, and GAN. Category B 
(\emph{High Novelty}) covers significant advances extending 
prior work such as BERT, ViT, and SAM. Category C 
(\emph{Moderate}) captures meaningful non-paradigmatic 
contributions. Category D (\emph{Incremental}) includes 
routine extensions and standard applications. Category E 
(\emph{Specious}) comprises fabricated or invalid claims 
including STAP Cells, Arsenic Life, and ICEP2. Dataset 
composition is summarized in Table~\ref{tab:dataset_composition}.

\begin{table}[h]
\centering
\caption{Evaluation dataset of 150 proposals across five 
novelty tiers with ground-truth score ranges and temporal 
evaluation windows.}
\label{tab:dataset_composition}
\small
\begin{tabular}{lccc}
\toprule
\textbf{Category} & \textbf{n} & \textbf{Ground Truth} & 
\textbf{Evaluation Window} \\
\midrule
A: Breakthrough & 30 & $8.5$--$10.0$ & 
2 years pre-publication \\
B: High Novelty & 30 & $7.0$--$8.5$  & 
1 year pre-publication \\
C: Moderate     & 30 & $5.0$--$7.0$  & 
Publication year \\
D: Incremental  & 30 & $3.5$--$5.0$  & 
Publication year \\
E: Specious     & 30 & $1.0$--$3.5$  & 
Post-retraction \\
\midrule
Total & 150 & & \\
\bottomrule
\end{tabular}
\end{table}

\subsubsection{Per-Category Deliberation Statistics}

Table~\ref{tab:deliberation_stats} reports deliberation 
frequency, debate depth, and literature retrieval behavior 
across novelty categories. Deliberation is triggered most 
frequently for extreme categories (A: 76.7\%, E: 83.3\%), 
where initial critic disagreement is highest. Breakthrough 
proposals retrieve fewer references with minimal contradiction, 
consistent with genuine gaps in prior literature. Specious 
claims exhibit high contradiction rates ($8.7 \pm 2.9$ per 
proposal), validating the system's ability to distinguish 
absence of evidence from evidence of contradiction.

\begin{table}[h]
\centering
\caption{Debate frequency and literature retrieval statistics 
across five novelty categories.}
\label{tab:deliberation_stats}
\small
\begin{tabular}{lcccc}
\toprule
\textbf{Category} & \textbf{Debate} & \textbf{Rounds} & 
\textbf{Refs} & \textbf{Contradictions} \\
 & \textbf{Triggered} & \textbf{(Avg.)} & 
\textbf{Retrieved} & \textbf{Found} \\
\midrule
A: Breakthrough & 76.7\% & 2.43 & $4.4 \pm 1.8$ & 
$0.3 \pm 0.6$ \\
B: High Novelty & 63.3\% & 2.11 & $5.9 \pm 2.2$ & 
$1.2 \pm 1.1$ \\
C: Moderate     & 46.7\% & 1.64 & $8.0 \pm 2.8$ & 
$3.8 \pm 1.8$ \\
D: Incremental  & 40.0\% & 1.50 & $9.4 \pm 3.1$ & 
$6.2 \pm 2.3$ \\
E: Specious     & 83.3\% & 2.68 & $8.5 \pm 3.2$ & 
$8.7 \pm 2.9$ \\
\bottomrule
\end{tabular}
\end{table}

\subsubsection{Cross-Domain Validation and Error Analysis}

Bias analysis revealed a slight conservative 
tendency for breakthrough hypotheses (mean signed error 
$= -0.17$) and a small positive bias for moderately novel 
hypotheses ($+0.12$). Of 13 misclassifications (8.7\% error 
rate), 10 arose from boundary ambiguity between creative 
application and incremental extension, with errors 
concentrated near category boundaries rather than reflecting 
gross misjudgments.

\section{Empirical Hypothesis Validation}
\label{app:validation}

\subsection{Coding Agent: Extended Evaluation on KramaBench}
\label{app:attrition}

To assess generalizability beyond pathology-specific 
hypotheses, we benchmarked the Coding Agent on 
KramaBench~\cite{lai2026kramabench}, a public data-science 
benchmark spanning diverse tabular reasoning tasks. This 
setting is strictly harder than our internal evaluation 
because the agent must autonomously locate relevant files 
before writing and executing code. We compare two conditions 
across four model backends: a baseline where task-relevant 
files are pre-attached ($N=94$ tasks), and a full 
dataset-discovery workflow where the agent searches a raw 
dataset root autonomously ($N=106$ tasks).

Tables~\ref{tab:baseline} and~\ref{tab:discovery} report 
performance under each condition. Under the oracle-file 
baseline, execution success rates range from 61.7\% to 
78.7\%, establishing an upper bound for the full pipeline. 
Adding autonomous discovery reduces execution rates to 
29.2\%--39.6\%, with the dominant failure mode being failure 
to finalize a valid file set rather than failures in code 
generation or execution.

\begin{table}[h]
\centering
\caption{Coding Agent performance on KramaBench with 
oracle file selection ($N=94$ tasks). \textit{Score\%}: 
composite benchmark score. \textit{Exact\% (Reports)}: 
exact-match rate over all tasks. \textit{Exact\% (Executed)}: 
exact-match rate conditioned on executed tasks. 
\textit{Executed\%}: fraction of tasks where the script 
ran successfully.}
\label{tab:baseline}
\small
\setlength{\tabcolsep}{4pt}
\begin{tabular}{lcccccc}
\toprule
\textbf{Model} & \textbf{Score\%} & \textbf{Exact} & 
\textbf{Exact\%} & \textbf{Exact\%} & 
\textbf{Executed} & \textbf{Exec.\%} \\
& & & \textbf{(Reports)} & \textbf{(Executed)} & & \\
\midrule
gpt-5.3-codex      & 42.18 & 30/94 & 31.9\% & 40.5\% & 
74/94 & 78.7\% \\
gpt-5.4-2026-03-05 & 40.79 & 27/94 & 28.7\% & 39.1\% & 
69/94 & 73.4\% \\
gpt-5.4-mini       & 32.04 & 24/94 & 25.5\% & 40.0\% & 
60/94 & 63.8\% \\
gpt-5.4-nano       & 29.90 & 19/94 & 20.2\% & 32.8\% & 
58/94 & 61.7\% \\
\bottomrule
\end{tabular}
\end{table}

\begin{table}[h]
\centering
\caption{Coding Agent performance on KramaBench with 
autonomous dataset discovery ($N=106$ tasks, discovery 
step budget $=6$). The drop in \textit{Executed\%} 
relative to Table~\ref{tab:baseline} reflects tasks 
that fail at the discovery stage and never reach 
execution.}
\label{tab:discovery}
\small
\setlength{\tabcolsep}{4pt}
\begin{tabular}{lcccccc}
\toprule
\textbf{Model} & \textbf{Score\%} & \textbf{Exact} & 
\textbf{Exact\%} & \textbf{Exact\%} & 
\textbf{Executed} & \textbf{Exec.\%} \\
& & & \textbf{(Reports)} & \textbf{(Executed)} & & \\
\midrule
gpt-5.3-codex      & 18.28 & 13/106 & 12.3\% & 32.5\% & 
40/106 & 37.7\% \\
gpt-5.4-2026-03-05 & 15.76 & 10/106 &  9.4\% & 32.3\% & 
31/106 & 29.2\% \\
gpt-5.4-mini       & 12.89 & 10/106 &  9.4\% & 23.8\% & 
42/106 & 39.6\% \\
gpt-5.4-nano       & 12.96 & 10/106 &  9.4\% & 25.0\% & 
40/106 & 37.7\% \\
\bottomrule
\end{tabular}
\end{table}

Table~\ref{tab:attrition} decomposes performance across 
the three pipeline stages. Stronger models (codex, 
gpt-5.4-2026-03-05) fail primarily at the discovery stage, 
while lighter models (gpt-5.4-mini) advance further through 
discovery but incur higher execution-retry costs. This 
confirms that improving dataset discovery is the primary 
lever for advancing end-to-end validation performance.

\begin{table}[h]
\centering
\caption{Three-stage task attrition under autonomous 
dataset discovery ($N=106$ tasks). Each row traces 
task throughput across file selection, execution, and 
exact match stages.}
\label{tab:attrition}
\small
\setlength{\tabcolsep}{4pt}
\begin{tabular}{lccccccl}
\toprule
\textbf{Model} & \textbf{Selected} & \textbf{Sel.\%} & 
\textbf{Exec.} & \textbf{Exec.\%} & \textbf{Exact} & 
\textbf{Ex.\%} & \textbf{Failure} \\
\midrule
gpt-5.3-codex      & 42/106 & 39.6\% & 40/106 & 37.7\% & 
13/106 & 12.3\% & No files \\
gpt-5.4-2026-03-05 & 33/106 & 31.1\% & 31/106 & 29.2\% & 
10/106 &  9.4\% & No files \\
gpt-5.4-mini       & 67/106 & 63.2\% & 42/106 & 39.6\% & 
10/106 &  9.4\% & Script fail \\
gpt-5.4-nano       & 52/106 & 49.1\% & 40/106 & 37.7\% & 
10/106 &  9.4\% & No files \\
\bottomrule
\end{tabular}
\end{table}

\begin{figure}[h]
  \centering
  \includegraphics[width=\linewidth]{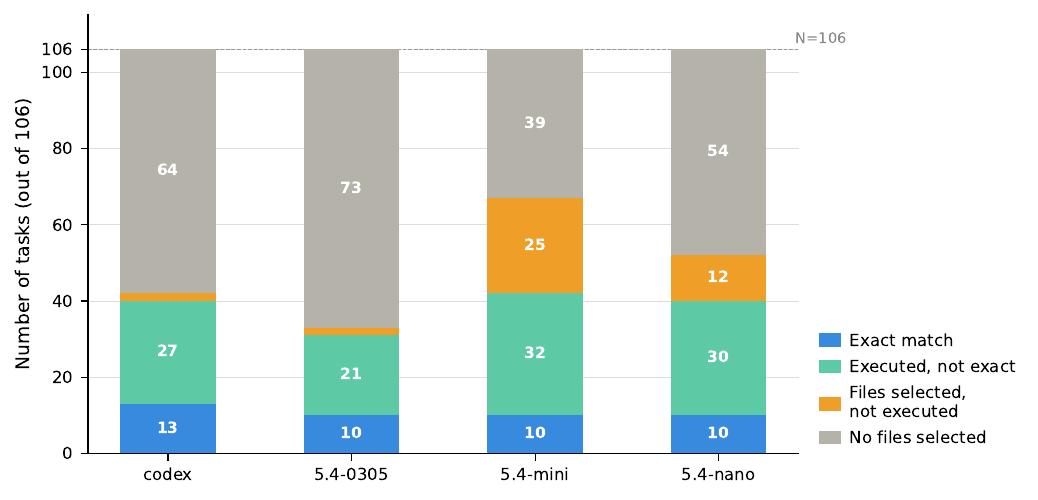}
  \caption{\textbf{Task attrition across three pipeline 
  stages on KramaBench ($N=106$ tasks per model).} Each 
  bar shows task distribution across four outcomes: exact 
  match, executed but not exact, files selected but not 
  executed, and no files selected. The dominant bottleneck 
  for stronger models is dataset discovery rather than 
  code generation.}
  \label{fig:attrition}
\end{figure}

\begin{figure}[h]
  \centering
  \includegraphics[width=\linewidth]{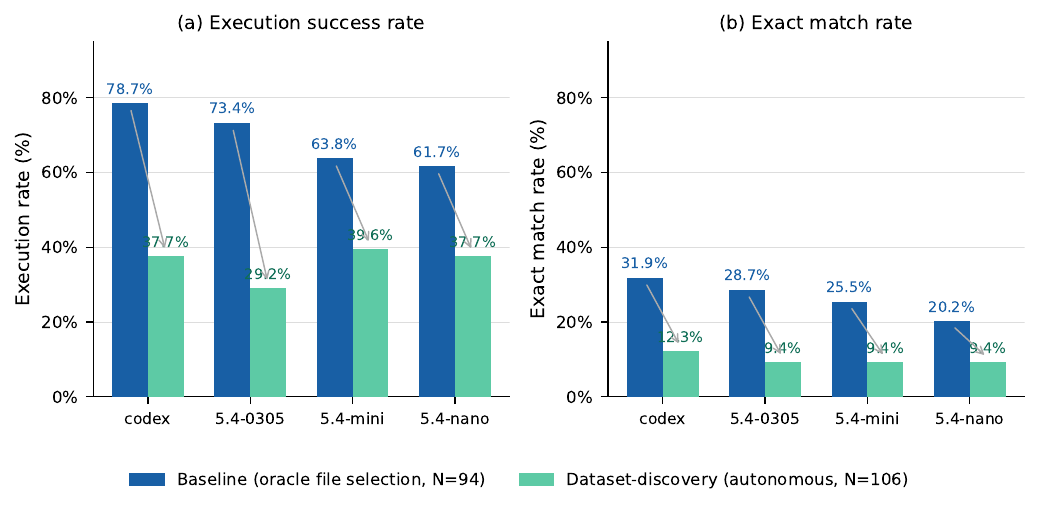}
  \caption{\textbf{Impact of dataset discovery on 
  execution and exact-match rates across four model 
  backends.} Dark blue: oracle file selection baseline 
  ($N=94$). Teal: autonomous discovery ($N=106$). 
  Execution rates fall by 24--44 percentage points under 
  discovery, confirming that code generation is performant 
  when supplied with correct inputs.}
  \label{fig:comparison}
\end{figure}

\subsection{Tool Registry and Orchestrator}
\label{app:tools}

The tool orchestration subsystem provides a unified control 
plane for interacting with heterogeneous clinical and 
computational tools through a plugin architecture. Each tool 
is registered through a structured specification file parsed 
into three progressively richer layers: a lightweight index 
for fast routing, an operational specification for planning, 
and a full description loaded on demand for execution. This 
layered design keeps the system token-efficient while scaling 
across a large tool catalog, and separates tool discovery, 
selection, planning, and execution into distinct concerns. 
New tools can be integrated without any backend code changes, 
making the subsystem fully modular and extensible.

The orchestrator interprets prompts from a user or autonomous 
agent, dynamically selects appropriate tools, and coordinates 
their execution over shared patient-level data and whole-slide 
images. Supported tool suites cover medical oncology workflows 
(survival analysis and treatment selection), histopathology 
pipelines (tumor, mitosis, tubule, collagen, and TIL 
segmentation, spatial transcriptomics, multiple instance 
learning, and foundation model inference), and emerging 
multimodal capabilities. Session-based 
state management persists tool outputs across analysis steps, 
enabling multi-step investigations without requiring users to 
re-specify intermediate results. A built-in tile server 
supports in-browser visualization of extracted masks and 
multi-channel overlays.

\begin{figure*}[t]
  \centering
  \includegraphics[width=\textwidth]{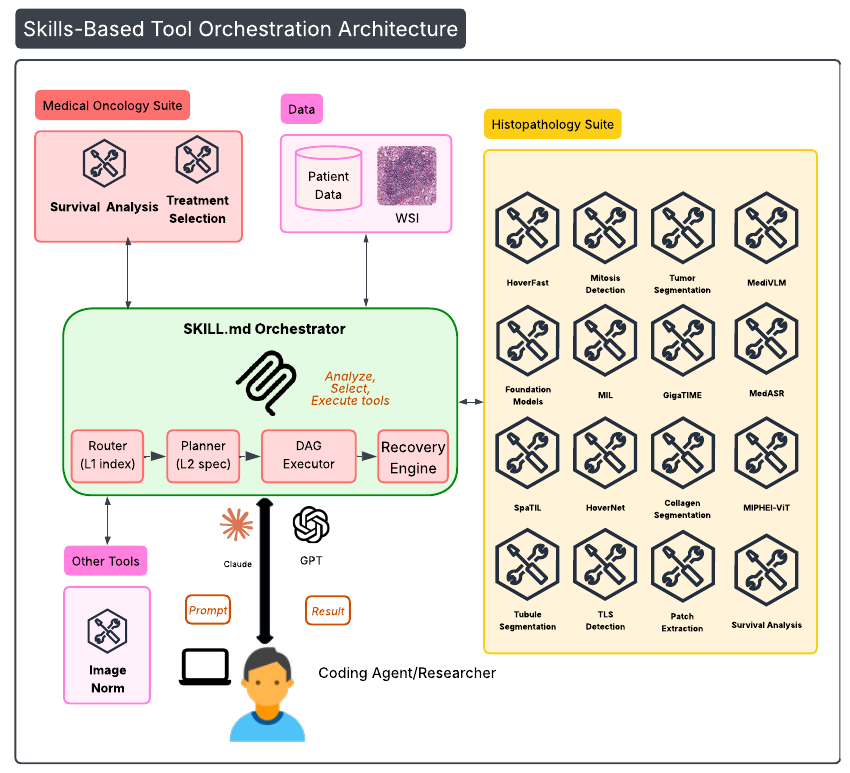}
  \caption{\textbf{Tool orchestration subsystem architecture.}
  The Orchestrator mediates between a coding agent or 
  researcher and distributed domain-specific tool suites by 
  analyzing prompts, selecting appropriate capabilities, and 
  coordinating execution over shared patient data and 
  whole-slide images.}
  \label{fig:sage_tools}
\end{figure*}

\begin{figure}[t]
  \centering
  \includegraphics[width=\linewidth]{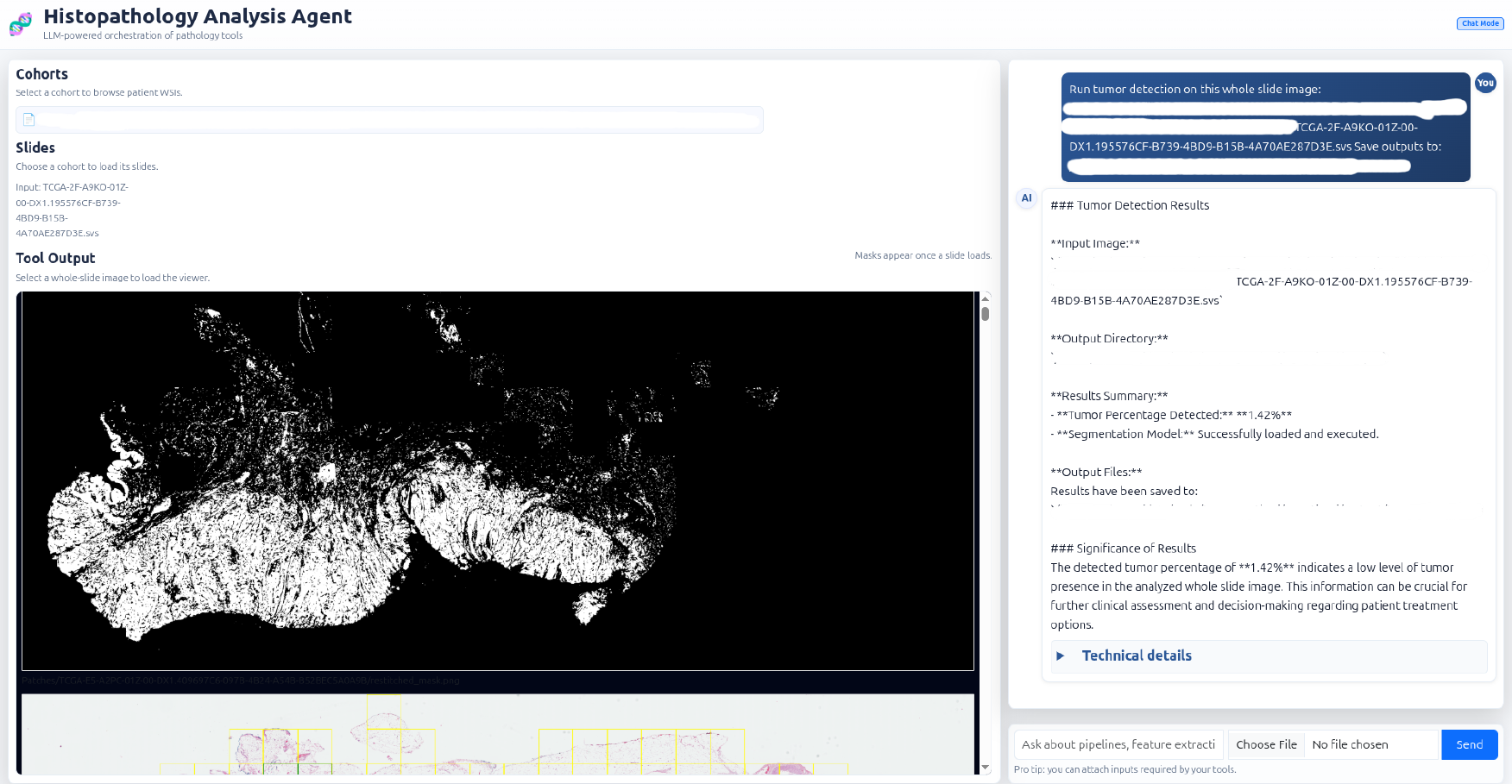}
  \caption{\textbf{Tumor detection via tool orchestration.}
  The subsystem responds to a tumor detection request by 
  invoking the tumor segmentation tool and returning a binary 
  mask highlighting detected tumor regions.}
  \label{fig:supp_tool_orchestration_mask}
\end{figure}

\begin{figure}[t]
  \centering
  \includegraphics[width=\linewidth]{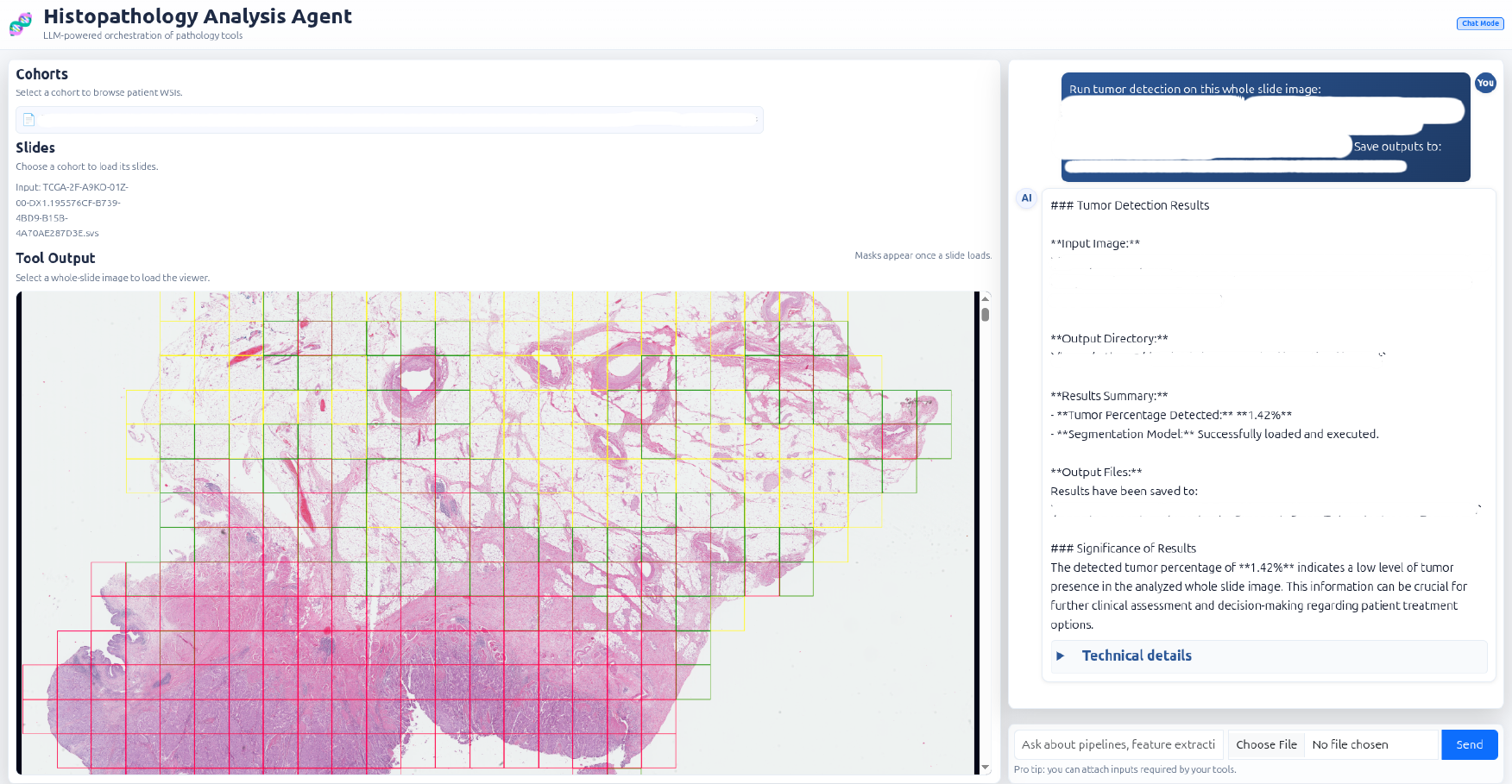}
  \caption{\textbf{Region-aware patch extraction via tool 
  orchestration.} Following tumor segmentation, the subsystem 
  composes segmentation and spatial reasoning tools to overlay 
  color-coded patches corresponding to tumor (red), 
  peritumoral (green), and remaining tissue (yellow) regions.}
  \label{fig:supp_tool_orchestration_patches}
\end{figure}

Figure~\ref{fig:sage_tools} provides an overview of the 
subsystem architecture. Figures~\ref{fig:supp_tool_orchestration_mask} 
and~\ref{fig:supp_tool_orchestration_patches} illustrate an 
example workflow in which a whole-slide image is uploaded, a 
tumor detection request is issued, and the subsystem invokes 
the appropriate segmentation tool to return a binary tumor 
mask followed by region-aware patch extraction across tumor, 
peritumoral, and remaining tissue regions.

\subsection{Expert Evaluation Protocol}
\label{app:expert_eval}

\paragraph{Expert panel composition.}
Four domain specialists independently evaluated hypotheses 
generated by each system: two board-certified urologists 
with experience in bladder cancer management and 
translational research, one computational pathologist 
specializing in whole-slide image analysis, and one 
molecular biologist with expertise in tumor immunology. 
Each expert was presented with hypotheses from their 
primary domain of expertise only, ensuring that scoring 
reflected genuine domain knowledge rather than 
generalist judgment.

\paragraph{Scoring rubric.}
\paragraph{Scoring rubric.}
Each hypothesis was evaluated using two criteria, \textit{Novelty} and \textit{Feasibility}, on a 1--10 ordinal scale, where scores of 1--4 indicate weak performance, 5--7 indicate moderate performance, and 8--10 indicate strong performance.

\textbf{Novelty} measures the originality of the hypothesis. 
Weak scores correspond to hypotheses based on well-known mechanisms or simple reformulations of existing ideas. 
Moderate scores indicate partial novelty or limited new insight. 
Strong scores indicate clearly new biological or clinical insights with meaningful conceptual innovation.

\textbf{Feasibility} measures how practical the hypothesis is to validate experimentally or computationally. 
Weak scores correspond to hypotheses with no clear validation pathway or unrealistic requirements. 
Moderate scores indicate hypotheses that are testable but may require substantial effort, resources, or additional tool development. 
Strong scores indicate hypotheses that can be readily tested using currently available datasets, models, or experimental protocols.

\paragraph{Evaluation procedure.}
Experts were blinded to the system identity that generated 
each hypothesis. 

\section{Experimental Results}
\label{app:results}


\subsection{CXCL13 as a Prognostic Biomarker in Muscle-Invasive 
Bladder Cancer}
\label{sec:supp_cxcl13}

\paragraph{Generated Hypothesis.}
In muscle-invasive bladder cancer (MIBC), higher CXCL13 
expression is associated with improved progression-free 
survival (PFS) and overall survival (OS). This hypothesis 
is grounded in the biological role of CXCL13 as a key 
mediator of tertiary lymphoid structure (TLS) formation: 
CXCL13 recruits B cells and T follicular helper cells, 
promoting immune niches that enhance antigen presentation 
and CD8$^+$ T cell activation, contributing to favorable 
clinical outcomes and immune checkpoint blockade response.

\paragraph{Validation Dataset and Design.}
The hypothesis was validated on the TCGA-BLCA cohort 
($N = 404$ patients, stage II--IV). Patients were 
stratified into CXCL13-high and CXCL13-low groups using 
the 75th percentile expression threshold, selected as the 
cutoff yielding statistically significant separation for 
both OS and PFS. Kaplan--Meier survival analysis with 
log-rank tests and multivariable Cox proportional hazards 
models adjusting for age, sex, and stage were used for 
validation.

\begin{figure}[h]
  \centering
  \includegraphics[width=\textwidth]{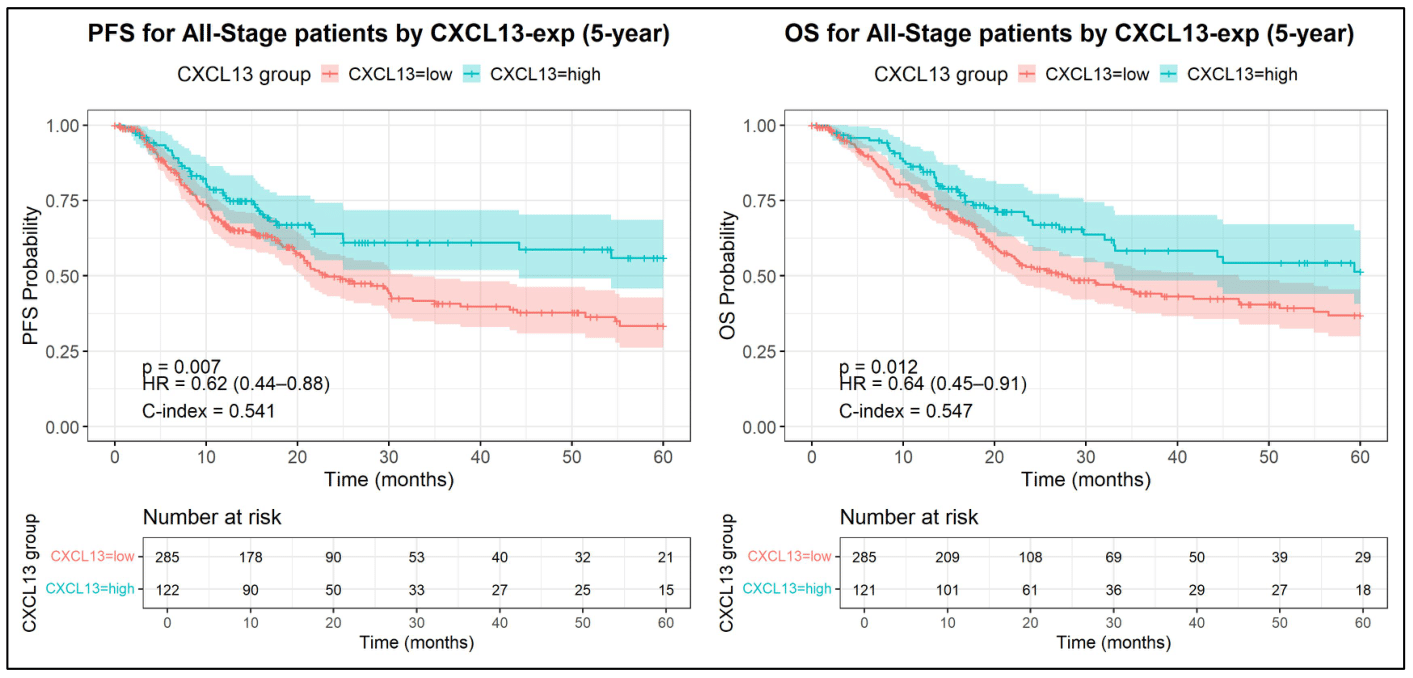}
  \caption{Kaplan--Meier survival curves for PFS (left) 
  and OS (right) in TCGA-BLCA stratified by CXCL13 
  expression. CXCL13-high patients exhibit significantly 
  longer survival under both endpoints.}
  \label{fig:cxcl13_os_pfs}
\end{figure}

\begin{figure}[h]
  \centering
  \includegraphics[width=\textwidth]{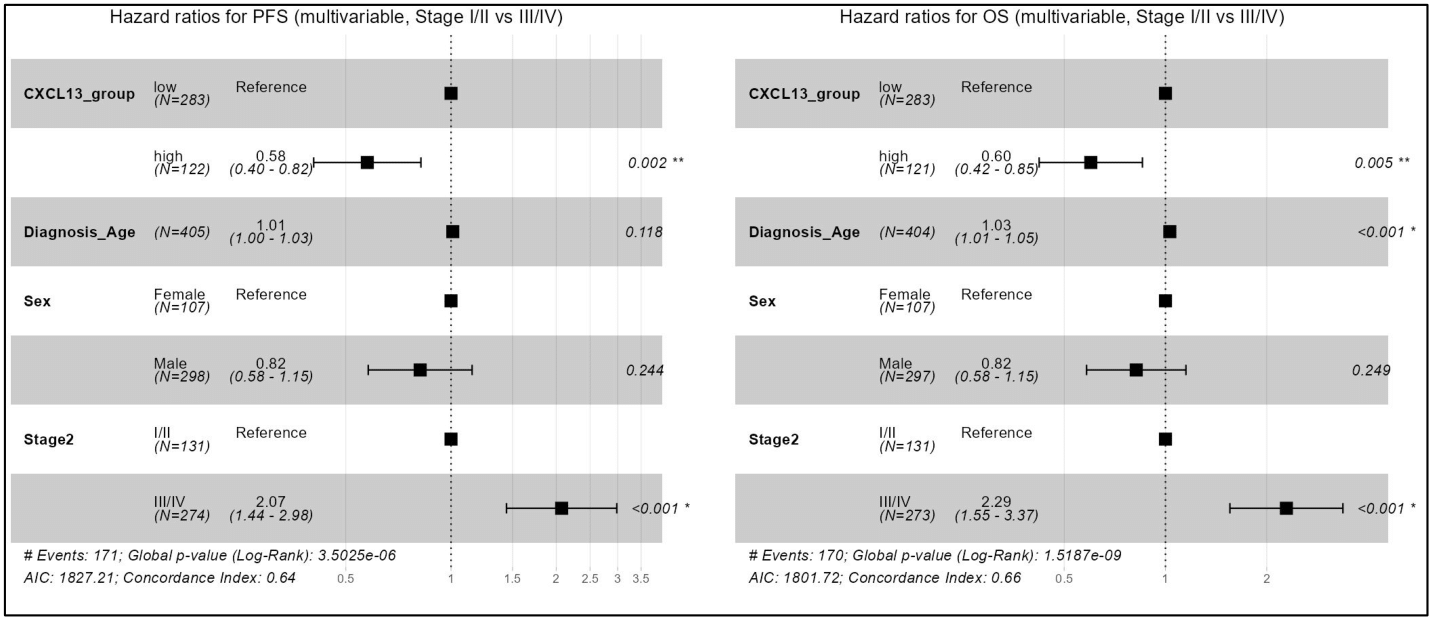}
  \caption{Forest plots from multivariable Cox models for 
  PFS and OS, adjusting for age, sex, and stage. CXCL13 
  remains independently associated with reduced hazard of 
  progression and death.}
  \label{fig:cxcl13_multivariable}
\end{figure}

\begin{figure}[h]
  \centering
  \includegraphics[width=\textwidth]{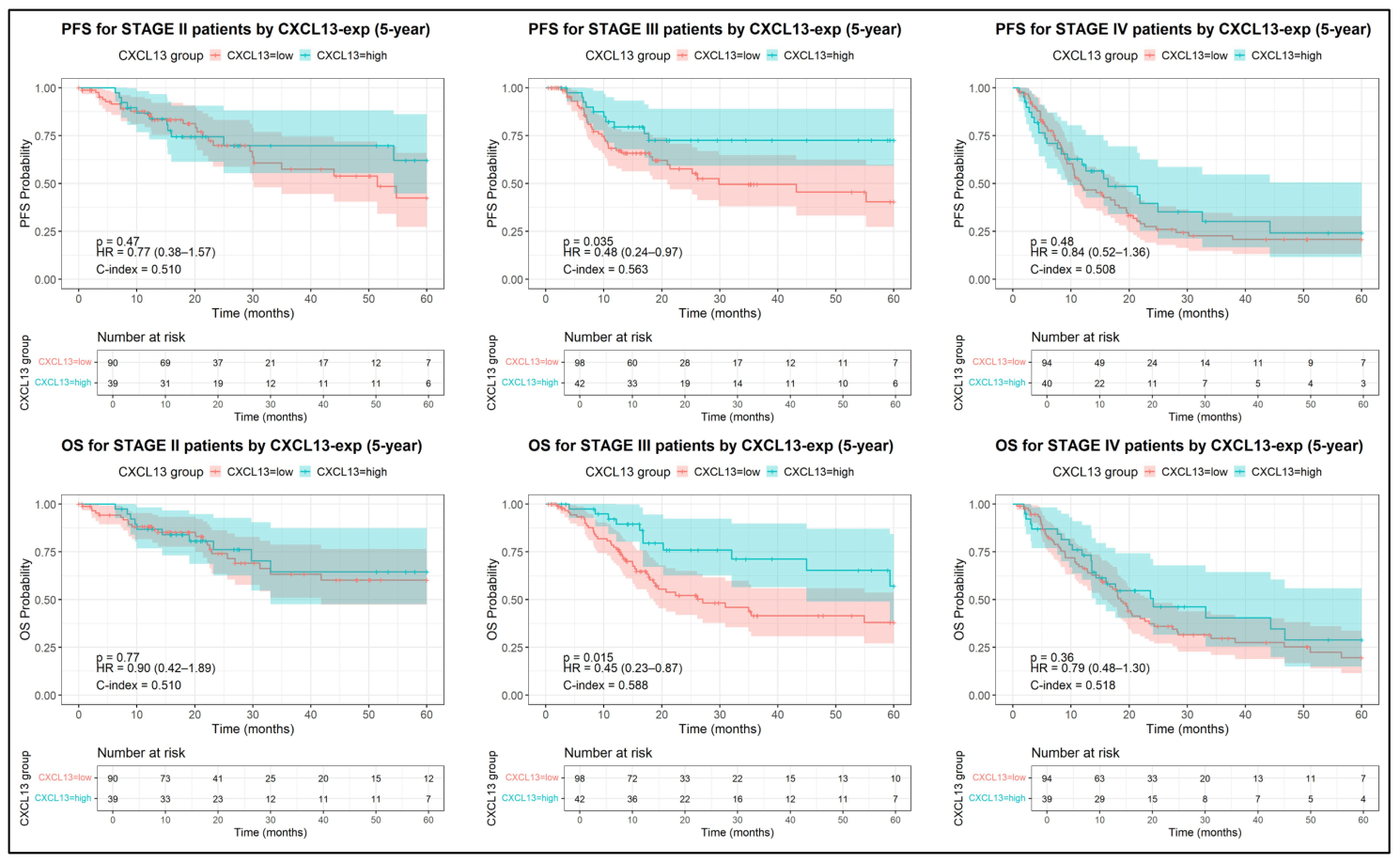}
  \caption{Stage-stratified Kaplan--Meier curves for PFS 
  (top) and OS (bottom) across Stage II, III, and IV 
  patients. The prognostic effect of CXCL13 is most 
  pronounced in Stage III disease.}
  \label{fig:cxcl13_subanalysis}
\end{figure}

\paragraph{Results.}
Kaplan--Meier analysis demonstrated that CXCL13-high 
patients had significantly improved OS (log-rank $p = 
0.0059$) and PFS (log-rank $p = 0.035$) compared to 
CXCL13-low patients (Figure~\ref{fig:cxcl13_os_pfs}). 
In multivariable Cox models, CXCL13 remained independently 
associated with reduced hazard after adjustment for 
clinical covariates: OS hazard ratio 0.60 (95\% CI 
0.42--0.85, $p = 0.005$, C-index = 0.66) and PFS hazard 
ratio 0.58 (95\% CI 0.40--0.82, $p = 0.002$, C-index = 
0.64) (Figure~\ref{fig:cxcl13_multivariable}). 
Stage-stratified analysis revealed that the prognostic 
effect was strongest in Stage III patients (OS: $p = 
0.015$, HR = 0.45; PFS: $p = 0.035$, HR = 0.48), while 
Stage II and Stage IV patients showed no significant 
associations (Figure~\ref{fig:cxcl13_subanalysis}).

\subsection{LAG3 Tumor Density as a Predictive Biomarker 
of Treatment Response in Bladder Cancer}
\label{sec:supp_lag3}

\paragraph{Generated Hypothesis.}
LAG3 tumor density, quantified from whole-slide images as 
an imaging-derived feature, is associated with response to 
immune checkpoint blockade in bladder cancer. This hypothesis 
is grounded in the established role of LAG3 as an inhibitory 
immune checkpoint receptor expressed on exhausted T cells: 
lower LAG3 tumor density in the tumor microenvironment is 
expected to reflect reduced immune exhaustion and greater 
susceptibility to checkpoint inhibitor therapy.

\paragraph{Validation Dataset and Design.}
The hypothesis was validated on an institutional bladder 
cancer cohort ($N = 57$ patients) with paired LAG3 tumor 
density measurements and treatment response labels 
(response\_label: 0 = non-responder, 1 = responder). 
Two endpoints were evaluated: (i) a predictive endpoint 
assessing association between LAG3 tumor density and 
treatment response using a Mann--Whitney U test, and 
(ii) a prognostic endpoint evaluating overall survival 
(OS) stratified by LAG3 tumor density via Kaplan--Meier 
analysis on training ($N = 46$) and holdout ($N = 11$) sets.

\begin{figure}[h]
  \centering
  \includegraphics[width=0.7\linewidth]
  {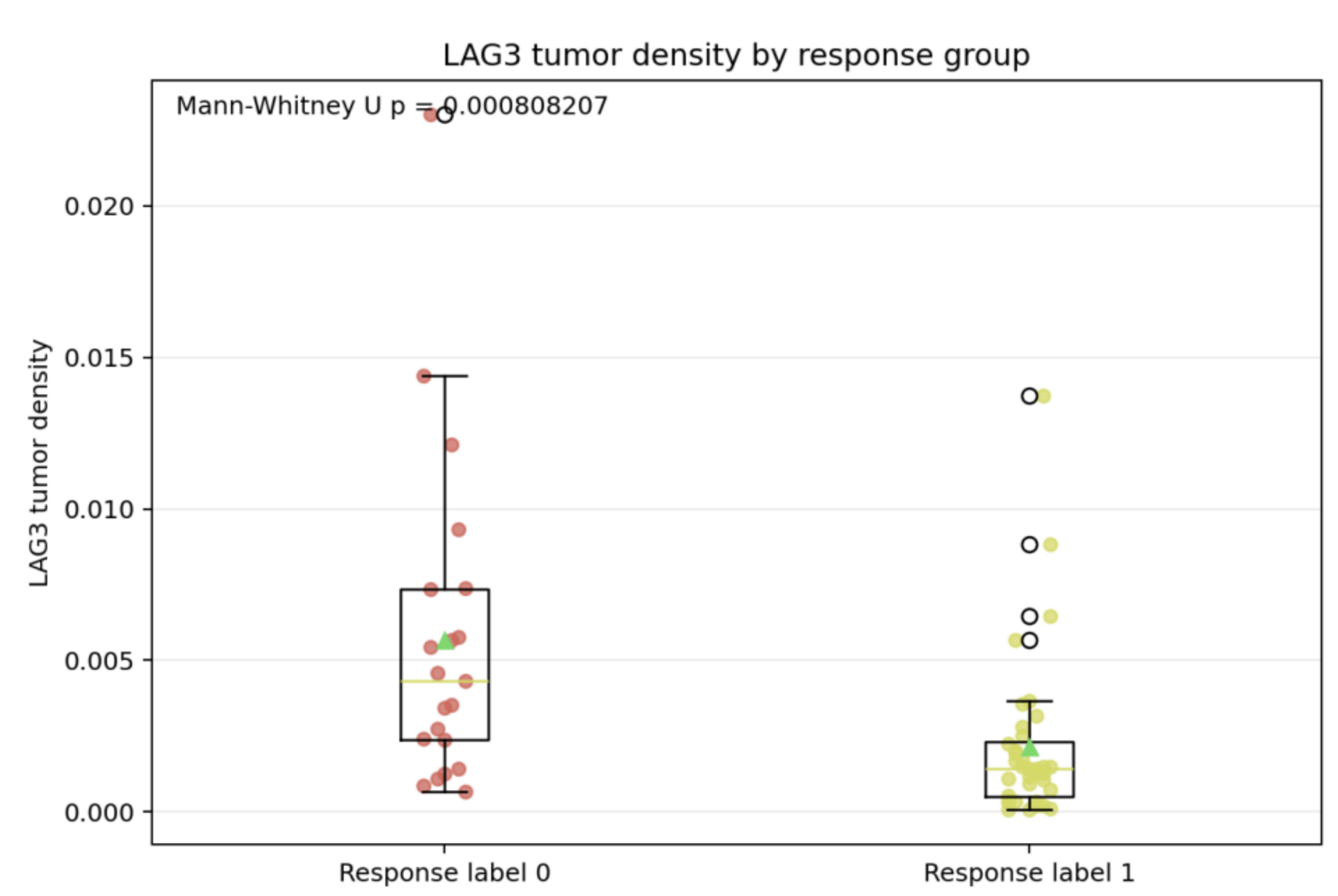}
  \caption{\textbf{LAG3 tumor density by treatment 
  response group.} LAG3 tumor density is significantly 
  lower in responders (response label 1) compared to 
  non-responders (response label 0), Mann--Whitney U 
  $p = 0.0008$. Each point represents a patient; 
  triangles indicate group means.}
  \label{fig:lag3_response}
\end{figure}

\paragraph{Results.}
\textbf{Predictive analysis.} LAG3 tumor density was 
significantly lower in responders (mean $= 0.00215$, 
median $= 0.00142$) compared to non-responders (mean 
$= 0.00567$, median $= 0.00432$), Mann--Whitney U 
$p = 0.0008$ (Figure~\ref{fig:lag3_response}), providing 
strong evidence supporting LAG3 tumor density as a 
predictive biomarker of treatment response.

\textbf{Prognostic analysis.} Survival analysis did not 
reveal statistically significant separation of OS by LAG3 
tumor density in either the training set (log-rank 
$p = 0.462$) or the holdout set (log-rank $p = 0.111$). 
The holdout set is small ($N = 11$, 10 events), making 
estimates unstable and the null result difficult to 
interpret definitively.

\paragraph{Interpretation.}
This example illustrates an important distinction that 
SAGE explicitly handles: predictive and prognostic 
biomarker claims are evaluated separately. LAG3 tumor 
density shows a strong association with treatment response 
but does not demonstrate independent prognostic value for 
overall survival in this cohort. This nuanced outcome 
reflects the transparent validation design of SAGE, which 
reports both supported and unsupported claims rather than 
selectively presenting positive results.

\newpage
\section*{NeurIPS Paper Checklist}

\begin{enumerate}

\item {\bf Claims}
    \item[] Question: Do the main claims made in the abstract and introduction accurately reflect the paper's contributions and scope?
    \item[] Answer:\answerYes{}
    \item[] Justification:  The abstract and introduction clearly state all four contributions: end-to-end biomarker discovery pipeline, multi-path ontological reasoning, debate-based novelty assessment, and biologically grounded validation. Each claim is directly supported by experimental results in the Results section.

\item {\bf Limitations}
    \item[] Question: Does the paper discuss the limitations of the work performed by the authors?
    \item[] Answer: \answerYes{}
    \item[] Justification: Limitations are discussed in the Conclusion

\item {\bf Theory assumptions and proofs}
    \item[] Question: For each theoretical result, does the paper provide the full set of assumptions and a complete (and correct) proof?
    \item[] Answer: \answerNA{}.
    \item[] Justification: The paper does not include theoretical results or formal proofs. All contributions are algorithmic and empirical. The path scoring metrics and ARG-based retrieval algorithm are defined formally but do not require proofs.

    \item {\bf Experimental result reproducibility}
    \item[] Question: Does the paper fully disclose all the information needed to reproduce the main experimental results of the paper to the extent that it affects the main claims and/or conclusions of the paper (regardless of whether the code and data are provided or not)?
    \item[] Answer: \answerYes{}
    \item[] Justification: The full pipeline architecture, agent roles, model assignments, and context allocation strategy are described in the Methods section and Appendix. Path scoring metrics are formally defined in Section on path generation and the corresponding Appendix. The novelty debate configuration, including disagreement threshold ($\sigma > 1.0$), convergence criterion ($\sigma < 0.5$), and maximum debate rounds (3), is specified in the Section on novelty agent, and the 150-proposal evaluation dataset is documented with temporal evaluation windows in Appendix. Survival analyses are conducted on TCGA-BLCA, a publicly available dataset accessible via cBioPortal, with all preprocessing steps and statistical methods specified. The KramaBench coding agent evaluation uses a public benchmark with model backends, discovery step budget (6), and execution retry budget (5) fully reported. The expert evaluation rubric and scoring procedure are provided in Appendix. Code is provided in the supplementary material.

\item {\bf Open access to data and code}
    \item[] Question: Does the paper provide open access to the data and code, with sufficient instructions to faithfully reproduce the main experimental results, as described in supplemental material?
    \item[] Answer: \answerYes{}.
    \item[] Justification: An anonymized code repository is provided in the supplementary material, containing the full SAGE pipeline implementation. The main validation experiments use TCGA-BLCA, publicly accessible via cBioPortal with no registration required. The KramaBench evaluation uses a public benchmark available at its original repository. The institutional cohort used in the LAG3 validation is available upon reasonable request to the corresponding institution. Instructions for data access, environment setup, and experiment reproduction are included in the supplementary material alongside the anonymized repository link.

\item {\bf Experimental setting/details}
    \item[] Question: Does the paper specify all the training and test details (e.g., data splits, hyperparameters, how they were chosen, type of optimizer) necessary to understand the results?
    \item[] Answer:\answerYes{}
    \item[] Justification: SAGE requires no model training, or 
optimizers as all agents use pre-trained LLMs via API. The computational 
pathology tools used in the validation subsystem are locked pre-trained 
models requiring no hyperparameter tuning, as described in Appendix. 
Agent model assignments and context sources are specified in Appendix. 
Path scoring weights and confidence thresholds are defined in the main text. 
Novelty debate hyperparameters, including disagreement threshold 
($\sigma > 1.0$), convergence criterion ($\sigma < 0.5$), and maximum 
debate rounds (3), are stated in the main text. For the main case study 
in the results Section, all 412 available TCGA-BLCA patients were used in 
evaluation mode. For the additional validated hypotheses in Appendix, 
dataset split details are specified per experiment. Coding agent execution 
budgets are reported, and expert evaluation criteria and 
scoring procedure are detailed in Appendix.

\item {\bf Experiment statistical significance}
    \item[] Question: Does the paper report error bars suitably and correctly defined or other appropriate information about the statistical significance of the experiments?
    \item[] Answer: \answerYes{}
    \item[] Justification: All tables report means with 95\% confidence 
    intervals and statistical significance where applicable. Survival 
    analyses report hazard ratios with 95\% CIs and log-rank p-values. 
    The novelty debate reports accuracy with significance markers 
    ($p < 0.05$). Diversity results include bootstrap CIs and permutation 
    test p-values.

\item {\bf Experiments compute resources}
    \item[] Question: For each experiment, does the paper provide sufficient information on the computer resources (type of compute workers, memory, time of execution) needed to reproduce the experiments?
    \item[] Answer: \answerYes{}
    \item[] Justification: All LLM-based experiments are conducted via 
    commercial API calls (OpenAI, Google, and Claude), requiring no local GPU 
    resources. Per-hypothesis running time and token costs are reported 
    in the results Section (Tables) and Appendix. Computational 
    pathology validation tools and survival analyses run on a single 
    NVIDIA A100 GPU. No additional compute beyond what is reported here and in 
    the paper was required.
    
\item {\bf Code of ethics}
    \item[] Question: Does the research conducted in the paper conform, in every respect, with the NeurIPS Code of Ethics \url{https://neurips.cc/public/EthicsGuidelines}?
    \item[] Answer: \answerYes{}
    \item[] Justification: All experiments use de-identified public datasets 
    or data collected under existing institutional oversight. Expert 
    evaluators participated voluntarily and are co-authors of this work. 
    The paper preserves anonymity in submission. No aspects of the research 
    deviate from the NeurIPS Code of Ethics.

\item {\bf Broader impacts}
    \item[] Question: Does the paper discuss both potential positive societal impacts and negative societal impacts of the work performed?
    \item[] Answer: \answerYes{}
    \item[] Justification: Positive impacts include accelerating biologically 
    interpretable biomarker discovery in underexplored cancers, reducing 
    reliance on expert intuition, and providing a domain-agnostic reasoning 
    pipeline that can be extended to other scientific domains by supplying 
    the appropriate domain-specific tool suite, as discussed in Section 5 
    (Conclusion). SAGE is designed as a research tool; any biomarker it 
    discovers requires independent validation across multiple cohorts before 
    clinical use, which is the responsibility of the end user. A potential 
    negative impact is that biomarker discovery may not generalize equitably 
    across demographic groups if the underlying knowledge graph reflects 
    historical disparities in research coverage.
    
\item {\bf Safeguards}
    \item[] Question: Does the paper describe safeguards that have been put in place for responsible release of data or models that have a high risk for misuse (e.g., pre-trained language models, image generators, or scraped datasets)?
    \item[] Answer: \answerNA{}.
    \item[] Justification: SAGE poses no significant misuse risk. It does 
    not release pre-trained models or scraped datasets. The knowledge graph 
    is constructed from publicly available peer-reviewed scientific 
    literature, and all LLM interactions use commercial APIs governed by 
    their respective providers' usage policies.

\item {\bf Licenses for existing assets}
    \item[] Question: Are the creators or original owners of assets (e.g., code, data, models), used in the paper, properly credited and are the license and terms of use explicitly mentioned and properly respected?
    \item[] Answer:\answerYes{}
    \item[] Justification: All datasets, models, and tools are properly 
    cited with their original publications. TCGA-BLCA is accessed via 
    cBioPortal under its open-access data use policy. KramaBench is used 
    under its original license. The SAGE codebase and knowledge graph are 
    released under the CC BY-ND 4.0 license, which permits sharing with 
    attribution while prohibiting modifications without permission.

\item {\bf New assets}
    \item[] Question: Are new assets introduced in the paper well documented and is the documentation provided alongside the assets?
    \item[] Answer: \answerYes{}.
    \item[] Justification: Two new assets are introduced: the SAGE knowledge 
    graph (41,053 nodes, 56,338 edges) documented in the main text and appendix, and the 150-proposal novelty evaluation dataset documented in Appendix. Both are released under CC BY-ND 4.0 alongside an 
    anonymized code repository provided in the supplementary material. 
    The expert evaluation questions per domain, scoring rubric, and 
    model-generated hypothesis outputs across all system configurations 
    are also included in the supplementary material as additional assets.

\item {\bf Crowdsourcing and research with human subjects}
    \item[] Question: For crowdsourcing experiments and research with human subjects, does the paper include the full text of instructions given to participants and screenshots, if applicable, as well as details about compensation (if any)? 
    \item[] Answer: \answerYes{}.
    \item[] Justification: Human evaluation was conducted by four domain 
    expert co-authors (two urologists, one pathologist, one molecular 
    biologist). The full scoring rubric, evaluation instructions, and 
    domain-specific questions are provided in the Appendix and the 
    supplementary material. No crowdsourcing platforms were used. 
    Experts participated as co-authors and were not separately compensated.

\item {\bf Institutional review board (IRB) approvals or equivalent for research with human subjects}
    \item[] Question: Does the paper describe potential risks incurred by study participants, whether such risks were disclosed to the subjects, and whether Institutional Review Board (IRB) approvals (or an equivalent approval/review based on the requirements of your country or institution) were obtained?
    \item[] Answer: \answerYes{}
    \item[] Justification: All patient data used in validation experiments 
    are either de-identified public datasets (TCGA-BLCA) or institutional 
    datasets held in the institutional repository, all of which are covered 
    by existing IRB approvals at the contributing institutions. Expert 
    co-authors evaluated computational hypotheses only and were not exposed 
    to identifiable patient data, posing no risk to human subjects. No new 
    patient data collection was performed for this study.

\item {\bf Declaration of LLM usage}
    \item[] Question: Does the paper describe the usage of LLMs if it is an important, original, or non-standard component of the core methods in this research? Note that if the LLM is used only for writing, editing, or formatting purposes and does \emph{not} impact the core methodology, scientific rigor, or originality of the research, declaration is not required.
    \item[] Answer: \answerYes{}
    \item[] Justification: LLMs are a core and original component of SAGE, 
    used across all pipeline stages including hypothesis generation, novelty 
    debate, explainability scoring, code generation, and results 
    interpretation. Model assignments per agent are fully described in 
    Appendix. LLMs were also used for writing assistance 
    in preparing this manuscript, but all scientific content, results, 
    and conclusions were verified by the authors.

\end{enumerate}

\end{document}